\documentclass[10pt,twocolumn,letterpaper]{article}

\usepackage[final]{cvpr}      

\usepackage{times}
\usepackage{epsfig}
\usepackage{graphicx}
\usepackage{amsmath}
\usepackage{amssymb}
\usepackage{multirow}
\usepackage{pifont}
\usepackage{booktabs}
\usepackage{epsfig}
\usepackage{booktabs,multirow}
\usepackage{graphics}
\usepackage{threeparttable}
\usepackage{color}
\usepackage[normalem]{ulem}
\usepackage{multirow}
\usepackage{float}
\usepackage{amsfonts}
\usepackage{bm}
\usepackage{enumitem}
\usepackage[numbers]{natbib}
\usepackage{array}
\usepackage{colortbl}
\definecolor{myy}{RGB}{126,95,0}
\definecolor{mygray}{gray}{.9}
\definecolor{bblue}{RGB}{30,80,120}
\definecolor{mygray1}{gray}{.7}
\usepackage{bm}
\usepackage{mathtools}
\usepackage{subcaption}
\usepackage[nopar]{lipsum}
\usepackage[export]{adjustbox}
\usepackage{bbold}

\definecolor{mygray}{RGB}{127,127,127}
\definecolor{mygreen}{RGB}{93,174,86}

\usepackage[dvipsnames]{xcolor}

\makeatletter
\newcommand{\thickhline}{%
	\noalign {\ifnum 0=`}\fi \hrule height 1pt
	\futurelet \reserved@a \@xhline
}
\makeatother

\newcommand\blfootnote[1]{%
  \begingroup
  \renewcommand\thefootnote{}\footnote{#1}%
  \addtocounter{footnote}{-1}%
  \endgroup
}

\usepackage{caption}
\usepackage[pagebackref=true,breaklinks=true,colorlinks,bookmarks=false]{hyperref}
\usepackage[utf8]{inputenc}

\usepackage{cleveref}
\crefname{section}{§}{§§}
\Crefname{section}{§}{§§}

\def\cvprPaperID{2981} 
\def\confName{CVPR}
\def\confYear{2023}

\begin{document}

\title{Referring Multi-Object Tracking}

\author{
Dongming Wu$^{1*\ddagger}$,
Wencheng Han$^{2*}$,
Tiancai Wang$^3$,
Xingping Dong$^{4}$,
Xiangyu Zhang$^{3,5}$,
Jianbing Shen$^{2\dagger}$\\
$^1$ Beijing Institute of Technology,
$^2$ SKL-IOTSC, CIS, University of Macau, 
$^3$ MEGVII Technology, \\
$^4$ Inception Institute of Artificial Intelligence, 
$^5$ Beijing Academy of Artificial Intelligence \\
{\tt\small \{wudongming97, wencheng256, shenjiangbingcg\}@gmail.com,}
{\tt\small wangtiancai@megvii.com} \\
 \href{https://github.com/wudongming97/RMOT}{https://github.com/wudongming97/RMOT}
}
\maketitle

\blfootnote{$*$Equal contribution. $\dagger$Corresponding author: \textit{Jianbing Shen}.
This work was supported in part by the FDCT grant SKL-IOTSC(UM)-2021-2023, the Grant MYRG-CRG2022-00013-IOTSC-ICI, and the Start-up Research Grant (SRG) of University of Macau (SRG2022-00023-IOTSC).
$\ddagger$The work was done during the internship at MEGVII Technology.
}

\begin{abstract} 

Existing referring understanding tasks tend to involve the detection of a single text-referred object. In this paper, we propose a new and general referring understanding task, termed referring multi-object tracking (RMOT). Its core idea is to employ a  language expression as a semantic cue to guide the prediction of multi-object tracking. To the best of our knowledge, it is the first work to achieve an arbitrary number of referent object predictions in videos. To push forward RMOT, we construct one benchmark with scalable expressions based on KITTI, named Refer-KITTI. Specifically, it provides 18 videos with 818 expressions, and each expression in a video is annotated with an average of 10.7 objects. Further, we develop a transformer-based architecture TransRMOT to tackle the new task in an online manner, which achieves impressive detection performance and outperforms other counterparts. 
\end{abstract}

\section{Introduction}
\label{sec:intro}
Recently, referring understanding~\cite{yu2016modeling,nagaraja2016modeling,khoreva2018video,talk2car}, integrating natural language processing into scene perception, has raised great attention in computer vision community. 
It aims to localize regions of interest in images or videos under the instruction of human language, which has many applications, such as video editing and autonomous driving.
For referring understanding, several significant benchmarks have been published.  Flickr30k~\cite{young2014image}, ReferIt~\cite{kazemzadeh2014referitgame}, and RefCOCO/+/g~\cite{yu2016modeling} have greatly encouraged the development of image-based referring tasks. More datasets (\eg, Lingual OTB99~\cite{li2017tracking}, Cityscapes-Ref~\cite{vasudevan2018object}, Talk2Car~\cite{talk2car}, Refer-DAVIS$_{17}$~\cite{khoreva2018video}, and Refer-Youtube-VOS~\cite{seo2020urvos}) are further proposed to cover the application in videos.

Despite these advanced progress, previous benchmarks have two typical limitations.
\textbf{First}, each expression tends to correspond to only one target. However, many objects have the same semantics in an open world, \ie, one single expression could refer to multiple objects. 
From this side, existing datasets lack flexible simulation on the multi-object scenarios, causing  referring understanding tasks far from satisfactory.
\textbf{Second}, the given expression may only describe part of frames for the video referring task, making the correspondence inaccurate. 
For example, given the expression `the car which is turning', we have to predict the overall trajectory even if the car has finished the turning action.
Obviously, a single expression cannot cover all short-term status of one target. 
Overall, existing datasets fail to provide an accurate evaluation under the situations of multiple referent targets and temporal status variances.


\begin{figure}
	\centering
	\includegraphics[width=\linewidth]{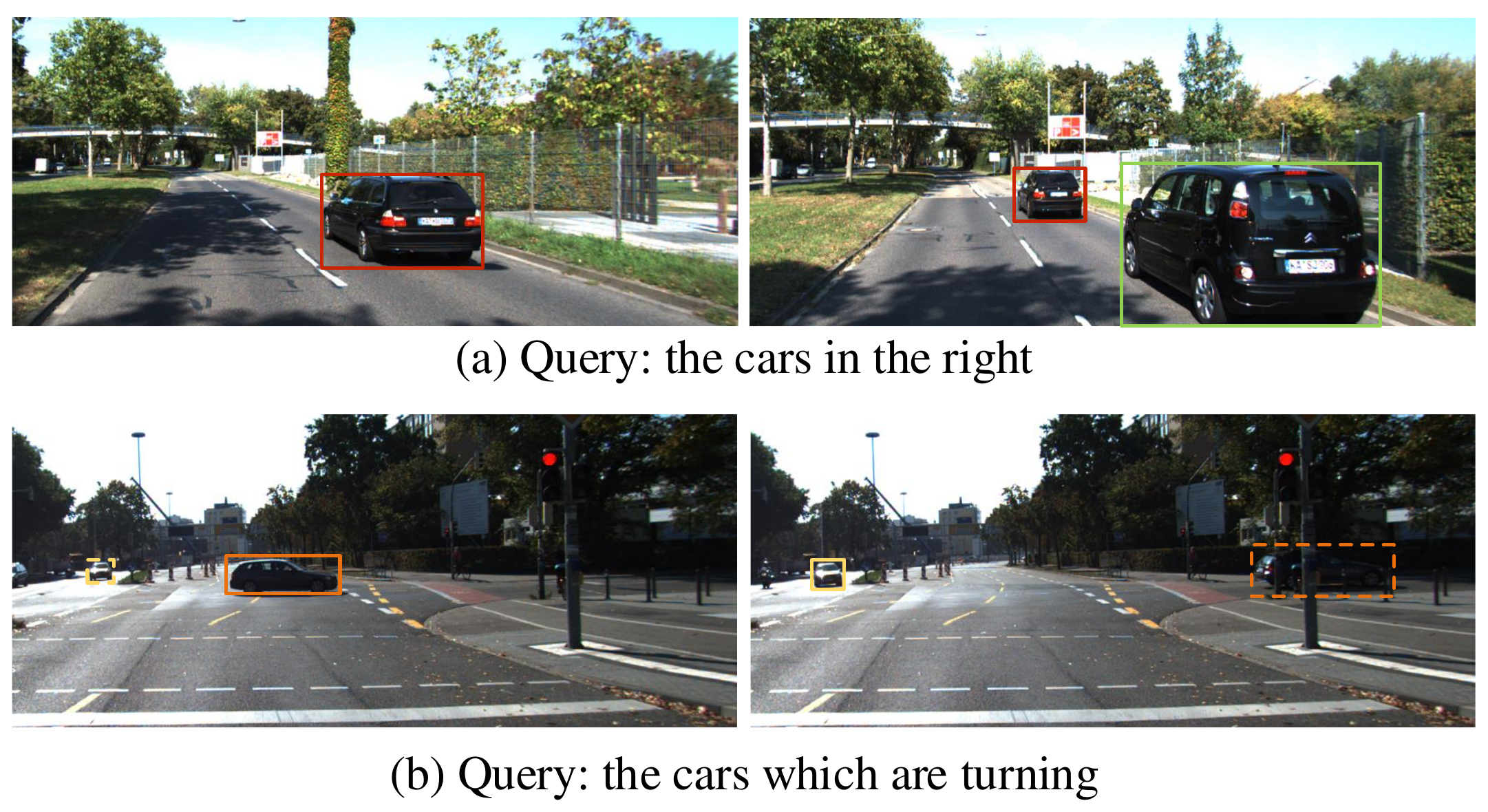}
	\vspace{-20pt}
	\caption{\textbf{Representative examples from RMOT}. The expression query can refer to multiple objects of interest (a),  and captures the short-term status with accurate labels (b).}
	\label{fig:motivation}
	\vspace{-10pt}
\end{figure}

 \begin{figure*}[t]
	\centering
	\resizebox{\textwidth}{!}
	{
		\includegraphics[width = 16cm]{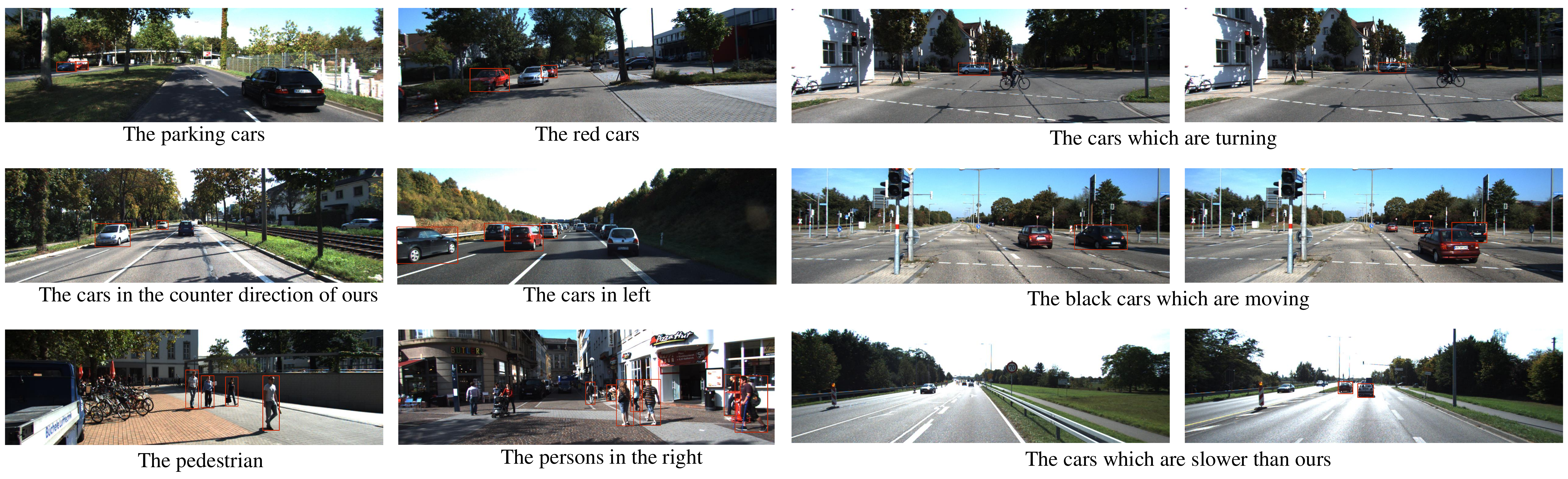}
	}
	\vspace{-22pt}
	\caption{\textbf{More examples of Refer-KITTI}.  It provides high-diversity scenes and high-quality annotations referred to by expressions. 	}
	\label{fig:dataset_examples}
	\vspace{-6pt}
\end{figure*}

To address these problems, we propose a novel video understanding task guided by the language description, named referring multi-object tracking (RMOT).
Given a language expression as a reference, it targets to ground all semantically matched objects in a video.
Unlike previous tasks, our proposed RMOT is much closer to the real environment, as each expression can involve multiple objects. 
For instance, the expression query `the cars in the right' corresponds to one object at the $20^{th}$ frame but two objects at the  $40^{th}$ frame (see Fig.~\ref{fig:motivation} (a)).
The phenomenon indicates that RMOT focuses more on finding the matched targets so that the referent number can be flexibly changed.
In addition, the temporal status variances are also considered in RMOT. 
As shown in Fig.~\ref{fig:motivation}(b), the given example shows the cars can be detected only when they start the turning action, and the tracking will be ended if they finish the activity.

To speed up the development of RMOT, we construct a new benchmark, \ie, Refer-KITTI, concerning the traffic scenes.
It is developed from the public KITTI~\cite{Geiger2012CVPR} dataset.
Compared to existing referring understanding datasets, it has three distinguishing characteristics: 
\textbf{i)} High flexibility with referent objects.
The number of objects described by each expression range from 0 to 105, with 10.7 on average.
\textbf{ii)} High temporal dynamics.
The temporal status of targets covers a longer time with more frames (varying in 0$\sim$400 frames), and the temporal variance of targets is accurately captured using our labeling tool.
\textbf{iii)} Low labeling cost with identification spread. We provide an effortless tool to annotate a target tracklet using only two clicks.

Although RMOT has a more flexible referring setting, it brings additional challenges: multi-object prediction and cross-frame association.
Towards this end, we propose an end-to-end differentiable framework for RMOT. Our model builds upon the recent DETR framework~\cite{carion2020end}, enhanced by powerful cross-modal reasoning and cross-frame conjunction.
It has an encoder-decoder architecture.
Specifically, we design an early-fusion module in the encoder to densely integrate visual and linguistic features, followed by a stack of deformable attention layers for further refining the cross-modal representations.
In the decoder, query-based embeddings interact with the cross-modal features to predict referent boxes. To track multi-objects, similar to MOTR~\cite{zeng2021motr}, we decouple the object queries into track query for tracking objects of previous frames and detect query for predicting the bounding boxes of new-born objects.


In summary, our contributions are three-fold. \textbf{First}, we propose a new task for referring multi-objects, called referring multi-object tracking (RMOT). 
It tackles limitations in the existing referring understanding tasks and provides multi-object and temporally status-variant circumstances.
\textbf{Second}, we formulate a new benchmark, Refer-KITTI, to help the community to explore this new field in depth. 
As far as we know,  it is the first dataset specializing in an arbitrary number of object predictions.
\textbf{Third}, we propose an end-to-end framework built upon Transformer, termed as TransRMOT. With powerful cross-modal learning, it provides impressive RMOT performance on Refer-KITTI compared to hand-crafted RMOT methods.


\begin{table}[t]
\centering
\small
\resizebox{0.48\textwidth}{!}{
	\setlength\tabcolsep{2pt}
	\begin{tabular}{l||cccc}
	\hline \thickhline
	 \cellcolor[gray]{0.9} & \cellcolor[gray]{0.9} &\cellcolor[gray]{0.9} &\cellcolor[gray]{0.9} & \cellcolor[gray]{0.9}\\
    \rowcolor[gray]{0.9}
	\multirow{-2}{*}{Dataset}  & \multirow{-2}{*}{Video}   & \multirow{-2}{*}{\shortstack{Images}}  & \multirow{-2}{*}{\shortstack{Instances\\per-expression}}  & \multirow{-2}{*}{\shortstack{Temporal ratio\\per-expression} }  \\
	\hline
	\hline
	RefCOCO~\cite{yu2016modeling}   & -  & 26,711     & 1    & 1    \\
	RefCOCO+~\cite{yu2016modeling}  & -  & 19,992     & 1    & 1    \\ 
	RefCOCOg~\cite{yu2016modeling}  & -  & 26,711      & 1    & 1    \\ 
	\hline
	Talk2Car~\cite{talk2car} &\checkmark    & 9,217       &    1  & - \\ 
	VID-Sentence~\cite{chen2019weakly}     &\checkmark    & 59,238       & 1  & 1    \\ 
 Refer-DAVIS$_{17}$~\cite{khoreva2018video}&\checkmark & 4,219 & 1 & 1\\
 Refer-YV~\cite{seo2020urvos}&\checkmark & 93,869 & 1 &1\\
 \hline
	Refer-KITTI                 			&\checkmark  & 6,650    &10.7   &  0.49\\ 
 \hline  \thickhline
	\end{tabular} }
\vspace{-2pt}
\caption{\textbf{Comparison of Refer-KITTI with existing datasets.} Refer-YV is short for Refer-Youtube-VOS. The temporal ratio represents the average ratio of referent frames covering the entire video sequence. `-' means unavailable.}
\label{table:dataset}
\vspace{-4mm}
\end{table}

\section{Related Work}

\begin{figure*}[t]
	\centering
	\resizebox{1\textwidth}{!}
	{
		\includegraphics[width = 16cm]{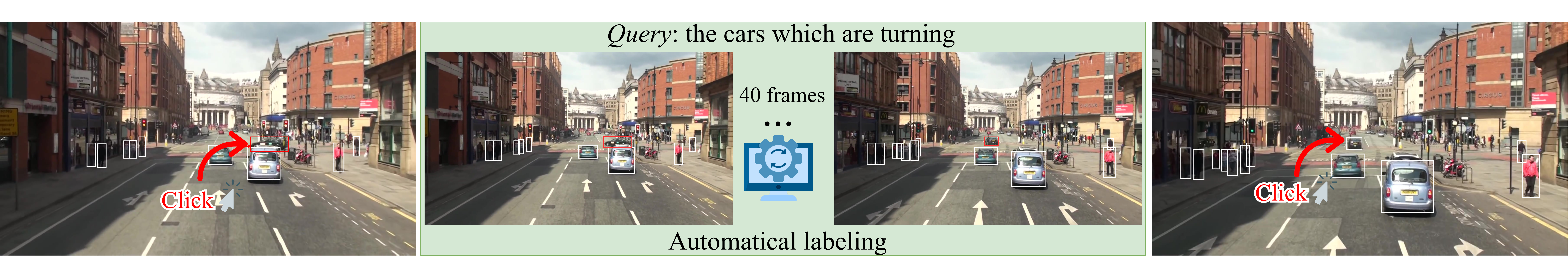}
	}
	\vspace{-20pt}
	\caption{\textbf{Labeling exemplar of our datasets.} The turning action is labeled with only two clicks on bounding boxes at the starting and ending frames. The intermediate frames are automatically and efficiently labeled with the help of unique identities.
	}
	\label{Fig:label}
	\vspace{-12pt}
\end{figure*} 

 \begin{figure}[t]
	\centering
	\resizebox{0.45\textwidth}{!}
	{
		\includegraphics[width = 16cm]{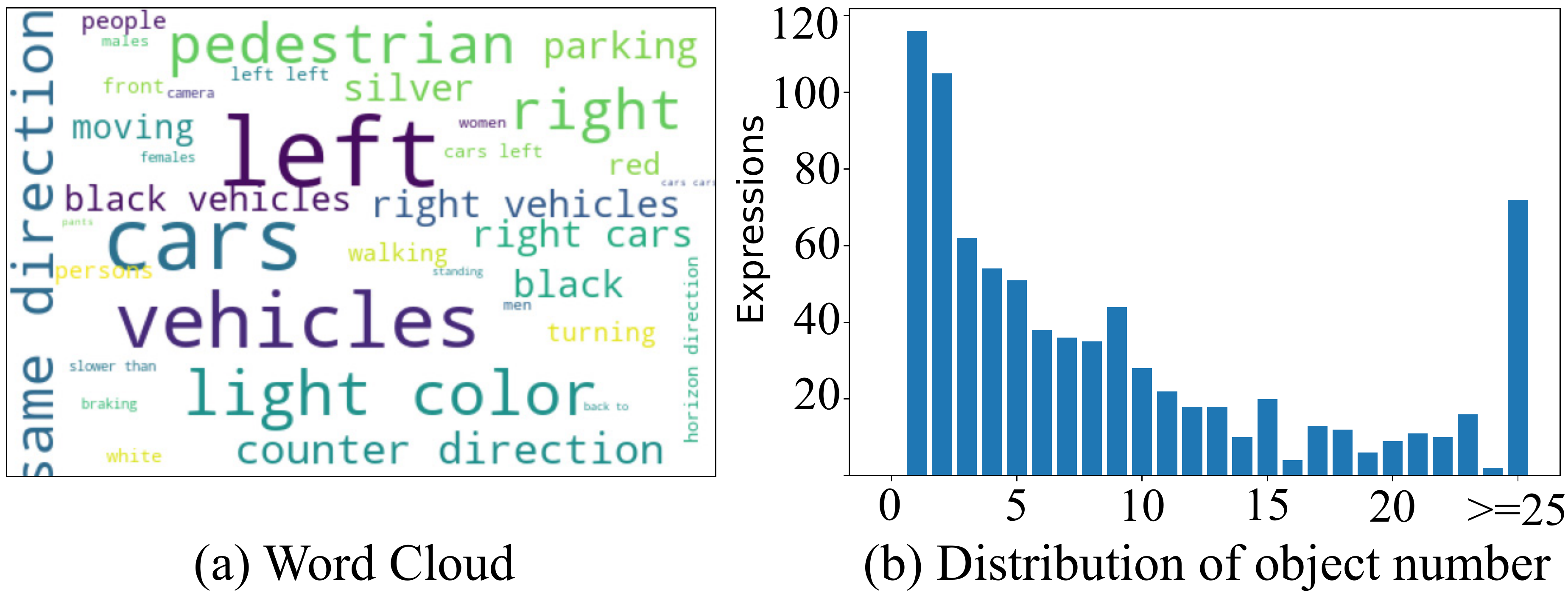}
	}
	\vspace{-7pt}
	\caption{\textbf{Statistics of Refer-KITTI} on (a) word cloud  and (b) distribution of object number per expression.
	}
	\label{fig:dataset_instance_histogram}
	\vspace{-4mm}
\end{figure}

\noindent\textbf{Referring Understanding Datasets.}
Many advanced datasets have greatly contributed to the progress of referring understanding.
Pioneering attempts (\eg, Flickr30k~\cite{young2014image}, ReferIt~\cite{kazemzadeh2014referitgame}, RefCOCO/+/g~\cite{yu2016modeling}) propose to employ a succinct yet unambiguous language expression to ground corresponding visual region within an image.
However, these datasets are fully image-based and do not fit well with common and practical video scenes.
Therefore, more efforts have been devoted to video-based benchmarks in recent years, such as Lingual OTB99~\cite{li2017tracking}, Cityscapes-Ref~\cite{vasudevan2018object}, VID-Sentence~\cite{chen2019weakly}, and Talk2Car~\cite{talk2car}.
In addition to grounding objects using bounding boxes, referring understanding is also involved in the video segmentation to formulate referring video object segmentation (RVOS). The mainstream of datasets include A2D-Sentences~\cite{gavrilyuk2018actor}, JHMDB-Sentences~\cite{gavrilyuk2018actor}, Refer-DAVIS$_{17}$~\cite{khoreva2018video}, Refer-Youtube-VOS~\cite{seo2020urvos}.
Although these datasets have undoubtedly promoted this field, they are still subject to two limitations, \ie, expressions referring to a single object and ignoring temporal variants. 
To alleviate them, we introduce RMOT, which targets multi-object and temporally-dynamic referring understanding. 
A thorough comparison between existing datasets and ours is summarized in Table~\ref{table:dataset}.
As seen, our proposed Refer-KITTI has a more flexible referent range (\ie, more boxes and fewer temporal ratios per expression). 
Although it has slightly fewer images than others, its performance stability can be guaranteed (see \S \ref{sec:quantitative_results}).

\noindent\textbf{Referring Understanding Methods.}
The core challenge of referring understanding is how to model the semantic alignment of cross-modal sources, \ie, vision and language.
Earlier algorithms mainly employed two separate stages~\cite{nagaraja2016modeling,mao2016generation,luo2017comprehension,yu2018mattnet,wang2019neighbourhood,liu2019learning,rufus2020cosine,khoreva2018video,sadhu2020video}: 1) the object detection stage that produces numerous object proposals using off-the-shelf detector; and 2) the object matching stage that learns the similarity between proposals and language expression to find the best-matched proposal as the final target. 
However, the performance of these methods relies heavily on the quality of the object detection. 
Later methods focus more on designing a one-stage pipeline~\cite{ye2019cross,liao2020real,luo2020multi,song2021co,tan2021look}. They fuse visual and linguistic modalities on early features instead of proposals, whereas the fusion strategies concentrate on employing a cross-modal attention mechanism.
Additionally, some works provide better semantic alignment interpretability via graph modeling~\cite{yang2019cross,yang2020graph}, progressive reasoning~\cite{huang2020referring,yang2021bottom,kesen2022modulating}, or multi-temporal-range learning~\cite{hui2021collaborative,ding2022language,wu2022multi}.
More recently, the Transformer-based models~\cite{mdetr,li2021referring,referformer,mttr,yang2022tubedetr} are becoming popular due to their powerful representation ability in cross-modal understanding.
Despite their progress, current referring understanding methods cannot process real-world and multi-object scenarios.
In contrast, our proposed Transformer-based model, TransRMOT, can deal well with these complex scenes.

 \begin{figure}[t]
	\centering
	\resizebox{0.48\textwidth}{!}
	{
		\includegraphics[width = 16cm]{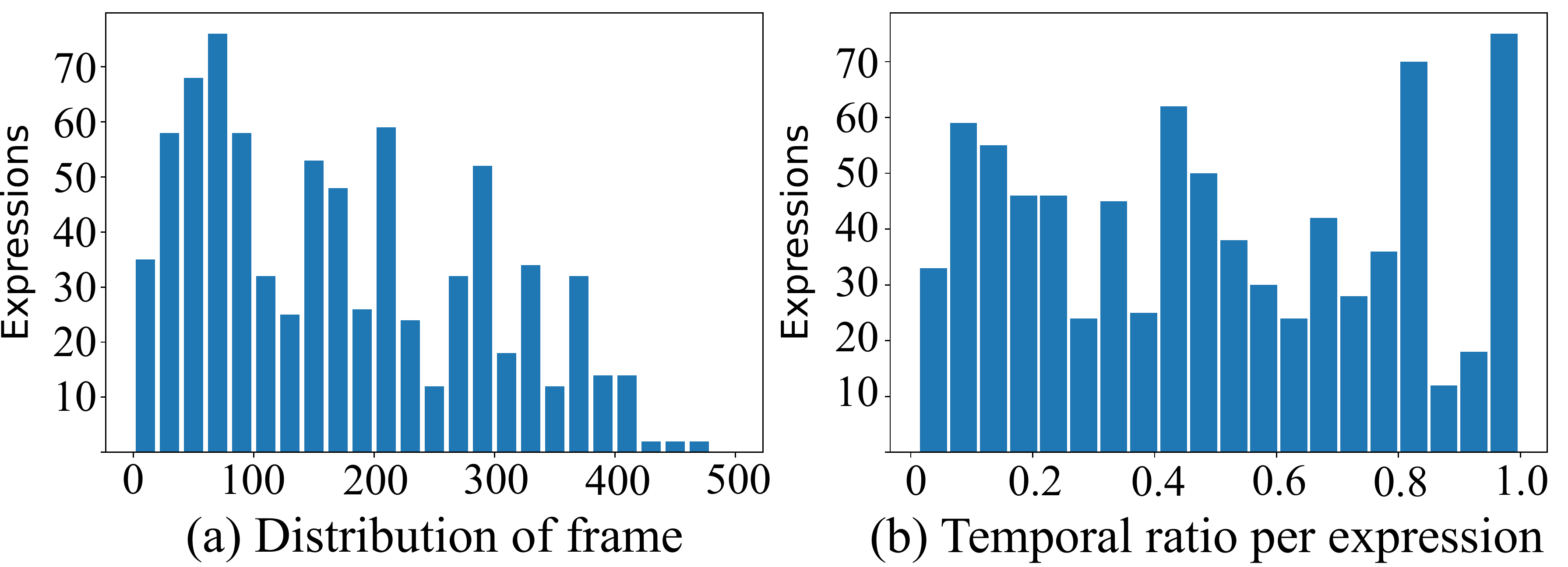}
	}
	\vspace{-20pt}
	\caption{\textbf{Temporal statistics.} (a) Distribution of  frame lengths. (b) Distribution of the ratio of referent frames covering  video. 
	}
	\label{fig:dataset_frame_histogram}
	\vspace{-4mm}
\end{figure}

\section{Benchmark}

To facilitate referring understanding, we construct the new dataset Refer-KITTI based on public KITTI~\cite{Geiger2012CVPR}.
There are two primary reasons why we choose it as our base. First,  KITTI contains various scenes, including pedestrian streets, public roads, highways, and so on, enriching the diversity of videos. Second, it provides a unique identification number for each instance, which helps annotators improve labeling efficiency using our proposed tool. 
We illustrate some representative examples in Fig.~\ref{fig:dataset_examples}.
 In the following,  we provide more details about Refer-KITTI.

\begin{figure*}[t]
	\centering
	\includegraphics[width=0.98\linewidth]{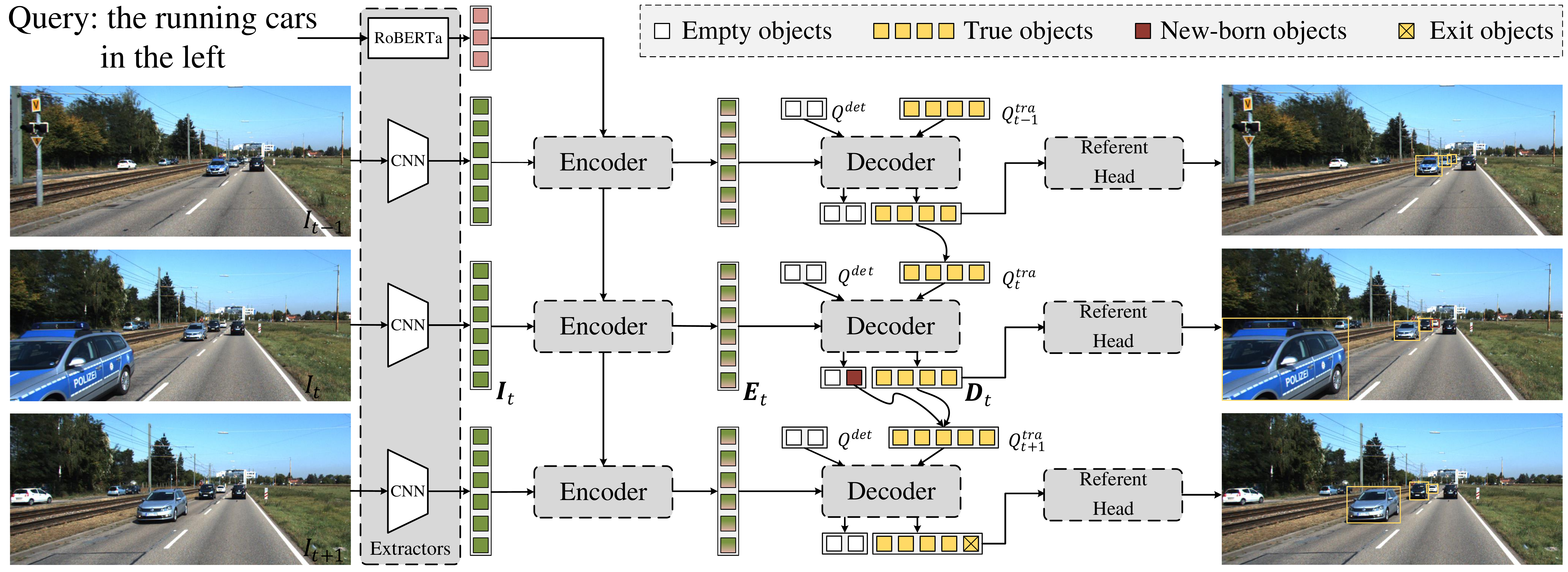}
	\vspace{-8pt}
	\caption{\textbf{The overall architecture of TransRMOT}. It is an online cross-modal tracker and includes four essential parts: feature extractors, cross-modal encoder, decoder, and referent head. 
 The feature extractors embed the input video and the corresponding language query into feature maps/vectors.
 The cross-modal encoder models comprehensive visual-linguistic representation via efficient fusion. The decoder takes the visual-linguistic features, detect queries and the track queries as inputs and updates the representation of queries. The updated queries are further used to predict the referred objects by the referent head.}
	\label{fig:method}
 \vspace{-2mm}
\end{figure*}

\subsection{Dataset Annotation with Low Human Cost}
\label{section:label}
In KITTI, each video has instance-level box annotations, and the same instance across frames has the same identification number.
We make full use of them and design an efficient labeling tool.
Fig.~\ref{Fig:label} shows an exemplar under the instruction of \textit{the cars which are turning}.  
To be specific, all bounding boxes are printed during annotation.
At the action-starting frame, we choose the turning car by clicking its box boundary.
Then we click the box boundary of the target again when it ends the turning action.
Instead of labeling the target objects frame by frame, our labeling tool can automatically propagate these labels to the intermediate frames according to their identification number.
Finally, the labeled information (\ie, frame ID, object ID, and box coordinates) and the corresponding expression are saved for training and testing.
The annotation procedure produces lower human costs than the frame-by-frame manner. 


\noindent\textbf{Collection Details.}
We assign each video at least three independent annotators and they review the entire video and manually create discriminative expressions to formulate our query sentences. Afterward, two other annotators carefully check the matching between expressions and video targets.

\noindent\textbf{Dataset Split.}
KITTI provides 21 high-resolution and long-temporal videos, but we abandon three over-complex videos and use the remaining 18  videos to formulate  Refer-KITTI.
We create a total of  818 expressions for Refer-KITTI using our labeling tool.
The word cloud of the expressions is shown in Fig.~\ref{fig:dataset_instance_histogram} (a).
Refer-KITTI is randomly split into separate \textit{train}, and \textit{test} sets, yielding a unique split consisting of 15 training videos and 3 testing videos. 

\subsection{Dataset Features and Statistics}
To offer deeper insights into Refer-KITTI, we next  discuss the discriminative features and descriptive statistics.

\noindent\textbf{High Flexibility with Referent Objects.} 
Different from previous datasets that contain just one referent object for each language expression, RMOT  is designed to involve an arbitrary number of predicted objects in videos.
Quantitatively, the expressions of Refer-KITTI mostly describe 0-25 objects, and the maximum number can be up to 105. 
On average, each expression in a video corresponds to\textbf{10.7} objects.
The per-expression object number distribution is shown in Fig.~\ref{fig:dataset_instance_histogram} (b).
These statistics are more representative of high-flexibility applications with referent objects.

\noindent\textbf{High Temporal Dynamic.}
Another real-world complexity is reflected in the temporal dimension of referent objects. 
Fig.~\ref{fig:dataset_frame_histogram} (a) shows the length distribution of frames per expression.
Most expressions of Refer-KITTI cover 0-400 frames, while the longest sequence has more than 600 frames.
Additionally, we show the per-expression temporal ratio covering the entire video in Fig.~\ref{fig:dataset_frame_histogram} (b).
It indicates many referent objects enter or exit from visible scenes.
The long time and undetermined ratio bring an additional challenge compared to existing works, \ie, \textit{ cross-frame object association}.

\subsection{Evaluation Metrics}
We adopt Higher Order Tracking Accuracy (HOTA)~\cite{luiten2020IJCV} as standard metrics to evaluate the new benchmark. 
Its core idea is calculating the similarity between the predicted and ground-truth tracklet.
Unlike MOT using HOTA to evaluate all visible objects, when those non-referent yet visible objects are predicted, they are viewed as false positives in our evaluation.
As the HOTA  score is obtained by combining  Detection Accuracy (DetA) and Association Accuracy (AssA), \ie, $\text{HOTA}\!=\!\sqrt{\text{DetA}\cdot \text{AssA}}$, it performs a great balance between measuring frame-level detection and temporal association performance. 
Here, DetA defines the detection IoU score, and AssA is the association IoU score.
In this work, the overall HOTA is calculated by averaging across different sentence queries.

\section{Method}


\subsection{Network Architecture}
The overall pipeline of our method is illustrated in Fig.~\ref{fig:method}. 
Taking the video stream as well as a language query as inputs, the goal is to output the track boxes of the corresponding query.
Similar to MOTR~\cite{zeng2021motr,zhang2022motrv2}, our model mainly follows the Deformable DETR~\cite{zhu2020deformable}, and we make several modifications on it to adapt the cross-modal inputs. It consists of four key components: feature extractors, cross-modal encoder, decoder and referent head.
The feature extractor first produces visual and linguistic features for the raw video and text.
Then, the cross-modal encoder fuses the features of two modalities.
Next, the decoder is used to update the representation of object queries. Finally, the referent head predicts the target sequences based on the predicted classification, bounding box and referent scores.

\noindent\textbf{Feature Extractor.}
Given a $T$-frame video, a CNN backbone model is used to extract the frame-wise pyramid feature maps, \eg, the $t^{th}$ frame for features $\bm{I}_t^l\in \mathbb{R}^{C_l \times H_l \times W_l}$, where $C_l, H_l, W_l$ represents the channel, height, width of the $l^{th}$ level feature map, respectively.
At the same time, we employ a pre-trained linguistic model to embed the text with $L$ words into 2D vectors $\bm{S}\!\in\!  \mathbb{R}^{L \times D}$, where $D$ is the feature dimension of word vectors.

\begin{figure}[t]
	\centering
	\includegraphics[width=0.8\linewidth]{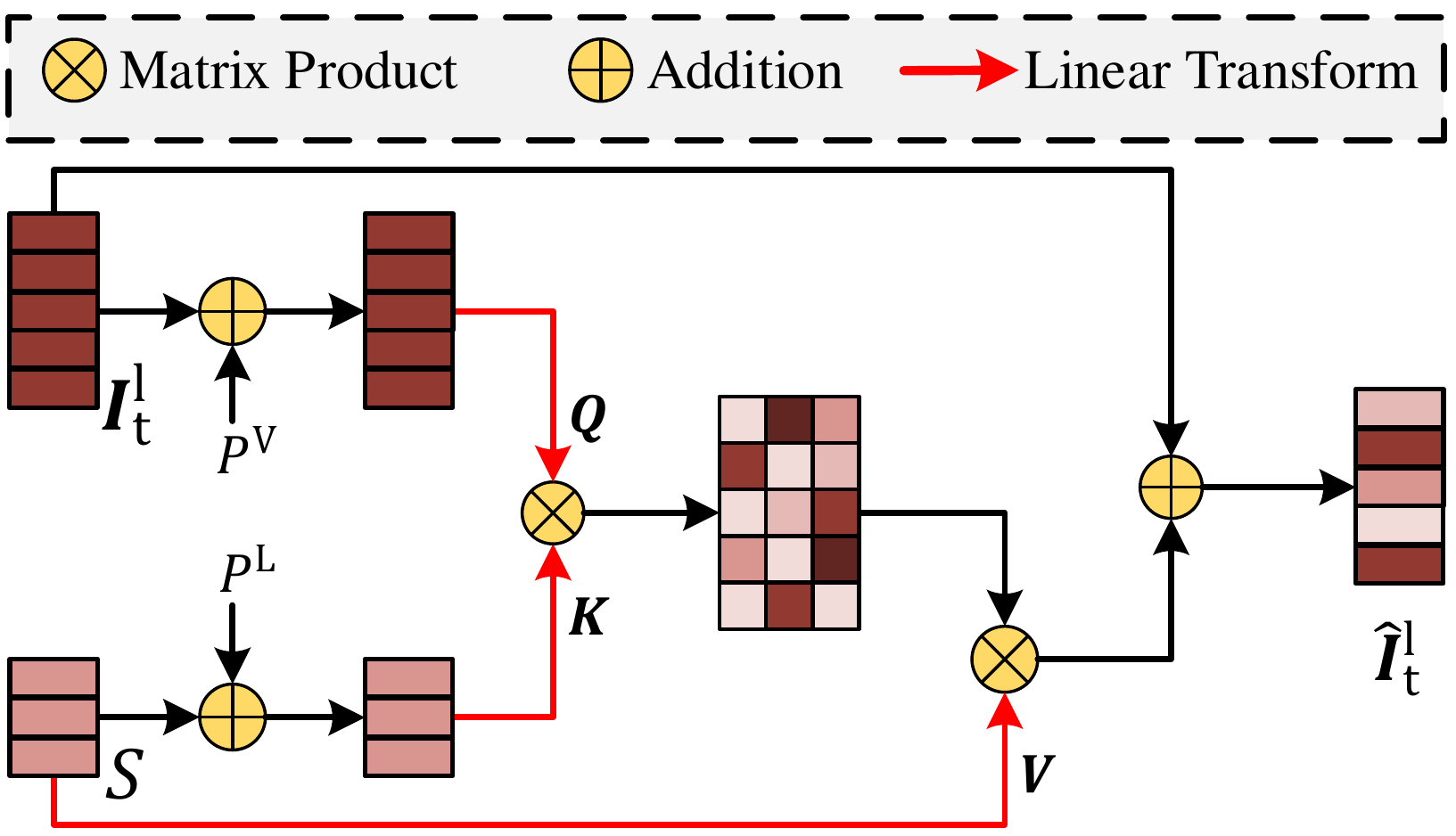}
	\vspace{-8pt}
	\caption{\textbf{The early-fusion module in our cross-modal encoder}. They achieve efficient visual-linguistic fusion followed by a stack of deformable encoder layers for further interaction.
	}
	\label{fig:encoder}
\end{figure}

\noindent\textbf{Cross-modal Encoder.}
The cross-modal encoder is responsible for accepting the visual and linguistic features and fusing them. 
The common strategy is to concatenate two kinds of features and feed them into the encoder to model dense connections via self-attention, like MDETR~\cite{mdetr}. 
However, the computation cost of self-attention is enormous due to the large token number of images.
To address this problem, we propose an early-fusion module to integrate the visual and linguistic features before deformable encoder layers. Our early-fusion module is illustrated in Fig.~\ref{fig:encoder}.

Specifically, given the $l^{th}$ level feature maps $\bm{I}_t^l$, we use a $1\!\times\! 1$ convolution to reduce its channel number to $d\!=\!256$, and flatten it into a 2D tensor $\bm{I}_t^l\!\in\! \mathbb{R}^{H_lW_l \times d}$. 
To keep the same channels with visual features, the linguistic features are projected into $\bm{S}\!\in\! \mathbb{R}^{L \times d}$ using a fully-connected layer.
Three independent full-connected layers transform the visual and linguistic features as $\bm{Q}$, $\bm{K}$, and $\bm{V}$:
\begin{equation}
	\small
	\vspace{-3pt}
	\begin{aligned}
	\bm{Q} &= \bm{W}_q (\bm{I}_t^l + P^V) \in \mathbb{R}^{H_lW_l \times d} , \\
	\bm{K} &= \bm{W}_k (\bm{S} + P^L)  \in \mathbb{R}^{L \times d},\\
	\bm{V} &= \bm{W}_v \bm{S}  \in \mathbb{R}^{L \times d},
	\end{aligned}
	\vspace{-1pt}
\end{equation}
where $\bm{W}$s are weights. $P^V$ and $P^L$  are position embedding of visual and linguistic features following~\cite{carion2020end,vaswani2017attention}.
We make matrix product on $\bm{K}$ and $ \bm{V}$,  and use the generated similarity matrix to weight linguistic features, \ie, $({\bm{Q}\bm{K}^\top}/{\sqrt{d}} )\bm{V} $. Here, $d$ is the feature dimension.
The original visual features are then added with the vision-conditioned linguistic features to produce the fused features $\hat{\bm{I}}_t^l$:
\begin{equation}
\small
	\vspace{-3pt}
	\hat{\bm{I}}_t^l = \frac{\bm{Q}\bm{K}^\top}{\sqrt{d}} \bm{V} + \bm{I}_t^l \in \mathbb{R}^{H_lW_l \times d}.
\end{equation}
After fusing two modalities, a stack of deformable encoder layers is used to promote cross-modal interaction:
\begin{equation}
\small
	\bm{E}_t^l = \text{DeformEnc}(\hat{\bm{I}}_t^l) \in \mathbb{R}^{H_lW_l \times d},
\end{equation}
where $\bm{E}_t^l$ is encoded cross-modal embedding, which will facilitate referring prediction in the following decoder.

\noindent\textbf{Decoder.}
The original decoder in the DETR framework uses learnable queries to probe encoded features for yielding instance embedding, further producing instance boxes and classes.
To associate objects between adjacent frames, we make full use of the decoder embedding from the last frame,  which is updated as  \textit{track query} of the current frame to track the same instance.
For new-born objects in the current frame, we adapt the original query from DETR, named \textit{detect query}. 
The tracking process is shown in Fig.~\ref{fig:method}.

Formally, let $\bm{D}_{t\!-\!1}\!\in\!\mathbb{R}^{N_{t\!-\!1}\!\times\! d}$ denote the  decoder embedding from the $(t\!-\!1)^{th}$ frame,  which is further transformed into \textit{track query} of  the $t^{th}$ frame, \ie, $Q^{tra}_{t}\!\in\!\mathbb{R}^{N_{t\!-\!1}'\!\times\! d}$, using self-attention and feed-forward network (FFN). 
Note that part of the $N_{t\!-\!1}$  decoder embeddings correspond to empty or exit objects, so we filter out them and only keep $N_{t\!-\!1}'$ true embeddings to generate  \textit{track query} $Q^{tra}_{t}$ in terms of their class score.
Let $Q^{det}\!\in\!\mathbb{R}^{N\!\times\! d}$ denote \textit{detect query}, which is randomly initialized for detecting new-born objects.
In practice, the two kinds of queries are concatenated together and fed into the decoder to learn target representation $\bm{D}_t$:
\begin{equation}
\small
	\bm{D}_t = \text{Decoder}(\bm{E}_t^l, \text{concat}(Q^{det}, Q^{tra}_t)) \in \mathbb{R}^{N_t\!\times\! d},
\end{equation}
where the number of output embedding is $N_t\!=\!N_{t\!-\!1}' + N$, including track objects and detect objects.

\noindent\textbf{Referent Head.} 
After a set of decoder layers, we add a referent head on top of the decoder. The referent head includes class, box and referring branches.
The class branch is a linear projection, which outputs a binary probability that indicates whether the output embedding represents a true or empty object. The box branch is a 3-layer feed-forward network with ReLU activation except for the last layer. It predicts the box location of all visible instances.
Another linear projection acts as the referring branch to produce referent scores with binary values. It refers to the likelihood of whether the instance matches the expression.

\begin{table*}[t]
	\centering
	\small
	\resizebox{0.9\textwidth}{!}{
		\setlength\tabcolsep{10pt}
		\begin{tabular}{l||cccccccccc| }
			\hline \thickhline
			\rowcolor[gray]{0.9}
			Method & HOTA ($\triangle$HOTA) & DetA & AssA & DetRe &	DetPr &	AssRe &	AssPr & LocA \\
			\hline
			\hline
			 FairMOT~\cite{zhang2021fairmot}                 &22.78($\pm$ 0.87)&14.43&39.11&16.44&45.48&43.05&71.65 & 74.77\\
			DeepSORT~\cite{Wojke2017simple}   &25.59($\pm$ 0.79)&19.76&34.31&26.38&36.93&39.55&61.05&71.34\\
			 ByteTrack ~\cite{zhang2022bytetrack}	       &24.95($\pm$ 0.84)&15.50&43.11&18.25&43.48&48.64&70.72&73.90\\
            CStrack~\cite{liang2022rethinking}              &27.91($\pm$ 0.73)&20.65	&39.10	&33.76	&32.61	&43.12	&71.82	&79.51\\
            TransTrack~\cite{sun2020transtrack}          &32.77($\pm$ 0.68)&23.31&45.71&32.33&42.23&49.99&78.74&79.48\\
            TrackFormer~\cite{meinhardt2022trackformer} & 33.26($\pm$ 0.65) & 25.44 & 45.87 & 35.21 & 42.19 & 50.26 & 78.92 & 79.63\\
			\hline 
			TransRMOT (Ours) &\textbf{35.54}($\pm$ 0.71)     &\textbf{28.25}&\textbf{46.25}&\textbf{39.22}&\textbf{45.94}&\textbf{50.69}&\textbf{80.67}&\textbf{79.79} \\
			\hline \thickhline
		\end{tabular} 
	}
	\vspace{-1pt}
	\caption{\textbf{Quantitative results on Refer-KITTI}. HOTA scores are reported, and $\triangle$HOTA presents score variance over three runnings.}
	\label{table:sota_kitti}
	\vspace{-4pt}
\end{table*}

\subsection{Instance Matching and Loss}

To train the model, we decouple the final loss as \textit{track loss} and \textit{detect loss}. As described before, our method predicts a flexible-size set of $N_t$ predictions for the $t^{th}$ frame,  including $N_{t\!-\!1}'$ \textit{tracking objects} and $N$ \textit{detection objects}. The $N_{t\!-\!1}'$  tracking objects and their ground-truth are one-to-one matched, while the detection objects are set predictions (\ie, the number of  predictions $N$ is larger than the number of true new-born objects).

Therefore, we first calculate the track loss using tracking prediction set $ \{{\bm{c}}^{tra}_{t,i}, {\bm{b}}^{tra}_{t,i}, {\bm{r}}^{tra}_{t,i}\}_{i=1}^{N_{t\!-\!1}'}$ and the ground-truth set 
$ \{\hat{\bm{c}}^{tra}_{t,i}, \hat{\bm{b}}^{tra}_{t,i}, \hat{\bm{r}}^{tra}_{t,i}\}_{i=1}^{N_{t\!-\!1}'}$ directly. Here, ${\bm{c}}^{tra}_{t,i}\!\in\!\mathbb{R}^1$ is a probability scalar indicating  whether this object is visible in the current frame.
${\bm{b}}^{tra}_{t,i}\!\in\!\mathbb{R}^4$ is a normalized vector that represents the center coordinates and relative height and width of the predicted box. 
${\bm{r}}^{tra}_{t,i}\!\in\!\mathbb{R}^1$ is a referring probability between the instance and the language description. The track loss $\mathcal{L}_t^{tra}$ is obtained via one-to-one computation:
\begin{equation}
\begin{aligned}
\small
\vspace{-10pt}
     \mathcal{L}_t^{tra} = \sum_{i=1}^{N_{t\!-\!1}'} \left[\lambda_{cls}  \mathcal{L}_{cls} ({\bm{c}}^{tra}_{t,i}, \hat{\bm{c}}^{tra}_{t,i} ) \right.
    +  \mathcal{L}_{box} ({\bm{b}}^{tra}_{t,i}, \hat{\bm{b}}^{tra}_{t,i})  \\
    +  \lambda_{ref}\mathcal{L}_{ref}({\bm{r}}^{tra}_{t,i}, \hat{\bm{r}}^{tra}_{t,i})\left.\right],
    \vspace{-3pt}
    \label{eq:track_loss}
\end{aligned}
\end{equation}
where $\mathcal{L}_{box}$ weights the L1 loss $\mathcal{L}_{L_1}$ and the generalized IoU loss $\mathcal{L}_{giou}$~\cite{giou}, \ie, $\mathcal{L}_{box}\!=\!\lambda_{L_1}\mathcal{L}_{L_1} + \lambda_{giou}\mathcal{L}_{giou}$. $\mathcal{L}_{cls}$ and $\mathcal{L}_{ref}$ are the focal loss~\cite{lin2017focal}.
$\lambda_{L_1}$, $\lambda_{giou}$, $\lambda_{cls}$, and $\lambda_{ref}$ are the corresponding weight coefficients.

Next, for detection objects, we need to find a bipartite graph matching which of the predicted objects fits the true new-born objects.
Let $\bm{y}^{det}_t\! =\!  \{{\bm{c}}^{det}_{t,i}, {\bm{b}}^{det}_{t,i}, {\bm{r}}^{det}_{t,i}\}_{i=1}^{N}$ denote detection set, and $\hat{\bm{y}}^{det}_t$ denote the new-born ground-truth. 
Then we search for a permutation of N predictions  $\delta \in P_n$  by minimizing matching cost:
\begin{equation}
\small
\vspace{-3pt}
	\hat{\delta} = \mathop{\arg \min} \limits_{\delta \in P_n} \mathcal{L}_{match}(\bm{y}^{det}_{t, \delta(i)}, \hat{\bm{y}}^{det}_{t}),
\end{equation}
where $\mathcal{L}_{match}\! =\! \mathcal{L}_{box}\! +\! \lambda_{cls} \mathcal{L}_{cls}$.
After obtaining the best permutation $\hat{\delta}$ with the lowest matching cost, we use it as a new index  of predictions $\{\bm{y}^{det}_{t, \hat{\delta}(i)}\}_{i=1}^N$  to compute the detect loss with ground-truth set $\hat{\bm{y}}^{det}_t$, as similar to Eq.~\ref{eq:track_loss}:
\begin{equation}
\begin{aligned}
\small
\vspace{-3pt}
     \mathcal{L}_t^{det} =& \sum_{i=1}^{N} \left[ \lambda_{cls}  \mathcal{L}_{cls} +\mathbb{1}  \mathcal{L}_{box}  \!+\! \mathbb{1}  \lambda_{ref}\mathcal{L}_{ref}\right],
\vspace{-3pt}
\end{aligned}
\end{equation}
where $\mathbb{1}$  refers to $\mathbb{1}_{\{\hat{\bm{c}}^{det}_{t,i} \neq \varnothing\}}$. 
Eventually, the final loss $\mathcal{L}^{final}$ is the summation of track loss and detect loss:
\begin{equation}
\small
\vspace{-5pt}
\mathcal{L}^{final} =  \sum_{t=1}^T( \mathcal{L}_t^{tra} + \mathcal{L}_t^{det}).
    \vspace{-2pt}
\end{equation}
As the first frame has no previous frames, its track query is set to empty. In other words, we only use the detect query to predict all new objects in the first frame.





\section{Experiments}

\subsection{Experimental Setup}
\noindent\textbf{Model Details.}
We adopt visual backbone ResNet-50~\cite{he2016deep} and text encoder RoBERTa~\cite{liu2019roberta} in our TransRMOT. 
Similar to Deformable DETR~\cite{zhu2020deformable}, the last three stage features $\{ \bm{I}^3_t, \bm{I}^4_t, \bm{I}^5_t\}$ from the visual backbone are used for further cross-modal learning. Besides, the lowest resolution feature map $\bm{I}^6_t$ is added via a $3\!\times\!3$ convolution with spatial stride 2 on the $\bm{I}^5_t$.
Each of the multi-scale feature maps is independently performed the cross-modal fusion. After that, deformable attention in the encoder and decoder integrates the multi-scale features.
The architecture and number of the encoder and decoder layer follow the setting of \cite{zhu2020deformable}.
The number of \textit{detect query} is set as $N\!=\!300$.

\begin{figure*}[t]
	\centering
	\includegraphics[width=\linewidth]{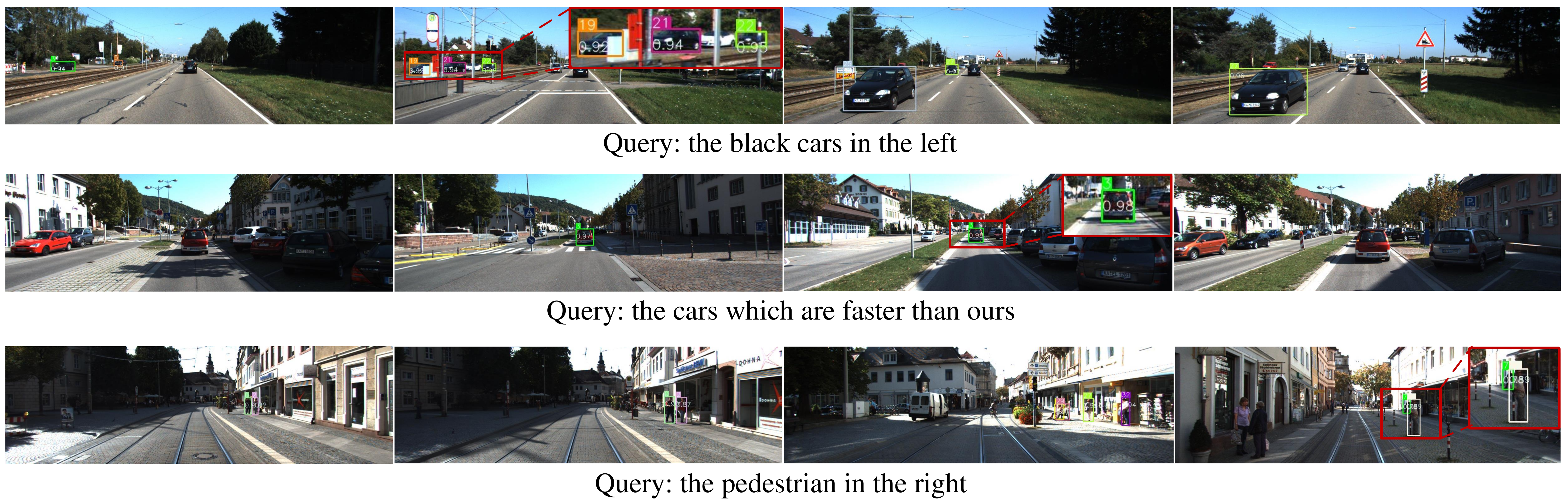}
	\vspace{-20pt}
	\caption{\textbf{Qualitative examples on Refer-KITTI}.  TransRMOT successfully predicts referent objects according to the given expression.
 }
	\label{fig:results}
	\vspace{-4pt}
\end{figure*} 

\begin{figure*}[t]
	\centering
	\includegraphics[width=\linewidth]{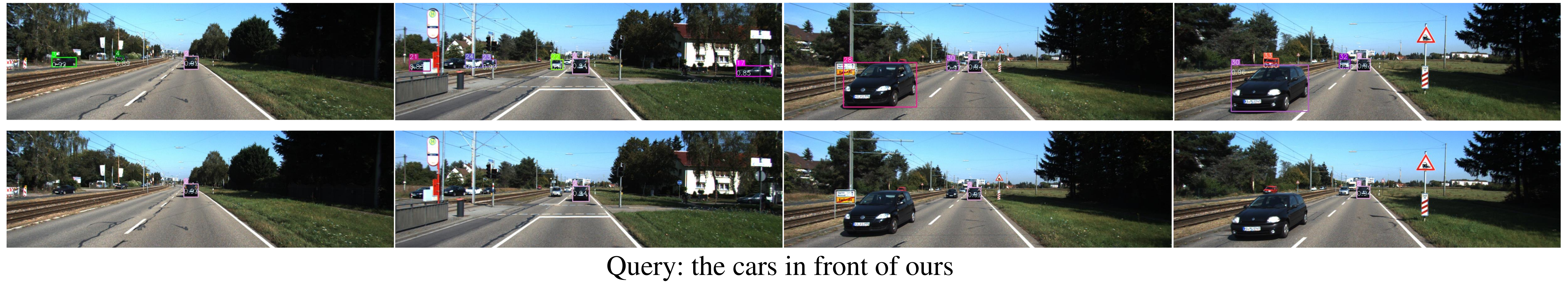}
	\vspace{-20pt}
	\caption{\textbf{Qualitative comparison between all visible objects (top) and the referent objects (bottom)}.   TransRMOT can capture all visible objects and highlight the referent ones. Please zoom in the figures for more details.
 }
	\label{fig:results_all}
	\vspace{-4pt}
\end{figure*} 

\noindent\textbf{Training.}
The parameters in the cross-modal module are randomly initialized for training, while the parameters in the text encoder are frozen during training.
The remained parameters are initialized from the official Deformable DETR weights~\cite{zhu2020deformable} pre-trained on the COCO dataset~\cite{coco}.
Random crop is used for data augmentation.
The shortest side ranges from 800 to 1536 for multi-scale learning.
Moreover, object erasing and inserting are added to simulate object exit and entrance following~\cite{zeng2021motr}. 
The loss coefficients are set as $\lambda_{cls}\!=\!5$, $\lambda_{L_1}\!=\!2$,  $\lambda_{giou}\!=\!2$, $\lambda_{ref}\!=\!2$.
AdamW optimizer is employed to train TransRMOT with base learning rate of 1e$^{-4}$. 
The learning rates of the backbone are set to 1e$^{-5}$. 
The model is trained for 100 epochs, and the learning rate decays by a factor of 10 at the 50$^{th}$ epoch. 
The overall training is deployed on 8 Nvidia 2080Ti GPUs with batch size of 1.

\noindent\textbf{Testing.}
TransRMOT is able to handle the arbitrary length of videos without post-process. 
At the  $t^{th}$ frame, it produces $N_t$ instance embeddings,  each corresponding to true or empty objects.
We choose these embeddings whose class score exceeds 0.7 to yield true object boxes.
Further, the final referent objects are determined from these true objects by a referring threshold $\beta_{ref}\!=\! 0.4$.

\subsection{Quantitative Results}
\label{sec:quantitative_results}
On top of Refer-KITTI, we examine the proposed TransRMOT and several competitors in Table~\ref{table:sota_kitti}.
Most previous approaches in referring understanding tasks are designed for single-object scenarios, which fail to predict boxes of multiple referent objects. 
Therefore,  we construct a series of CNN-based competitors by integrating our cross-modal fusion module into the detection part of multi-object tracking models, such as FairMOT~\cite{zhang2021fairmot}, DeepSORT~\cite{Wojke2017simple}, ByteTrack~\cite{zhang2022bytetrack},  and CStrack~\cite{liang2022rethinking}. 
These competitors follow a \textit{tracking-by-detection} paradigm and employ independent trackers to associate each referent box.
More model details can be found in \textit{supplementary materials}.
From Table~\ref{table:sota_kitti}, we can see that our TransRMOT outperforms other CNN counterparts by a large margin.

In addition, we compare with Transformer-based works, such as TransTrack~\cite{sun2020transtrack} and TrackFormer~\cite{meinhardt2022trackformer}, by adding the cross-modal learning parts. 
As shown in Table~\ref{table:sota_kitti}, both two methods perform worse than our TransRMOT across all metrics but achieve better scores than CNN-based models. 
Overall,  these experiments indicate the model priority of our proposed TransRMOT.
Moreover, the stability of Refer-KITTI is also evaluated by training all models three times with different seeds.
The slight performance variance (\ie, $\triangle$HOTA$<$0.87) shows great benchmark stability.


 \begin{table*}[t!]
 \small
    \begin{center}
    \ \ 
    \setlength{\tabcolsep}{5pt}
    \begin{subtable}[t]{0.48\linewidth}
        \begin{tabular}{p{2.6cm}|ccccc}
        \hline\thickhline
        \rowcolor[gray]{0.9}
        Fusion Way & HOTA & DetA & AssA & DetRe & DetPr\\
        \hline
      \textit{w/o} fusion 	&17.01 &19.05&15.26&22.00&56.07  \\
	Concatenation 			&28.61 &22.48&37.69&27.16&51.16 \\
        Language as query 		&33.29 &26.15&43.80&35.94&44.83\\
        Ours                                    & \textbf{35.54}&\textbf{28.25}&\textbf{46.25}&\textbf{39.22}&\textbf{45.94}\\
        \hline
        \end{tabular}
        \vspace{-1mm}
        \caption{Comparison on cross-modal fusion way. }
    \end{subtable}
    \vspace{2mm}
    \  \   \   \  
     \setlength{\tabcolsep}{5pt}
    \begin{subtable}[t]{0.48\linewidth}
        \begin{tabular}{p{2.25cm}|ccccc}
        \hline\thickhline
        \rowcolor[gray]{0.9}
        Association Way & HOTA & DetA & AssA & AssRe & AssPr\\
        \hline
        \textit{w/o} track query &- &	26.95&	- &	-	& -\\
	SORT~\cite{Bewley2016_sort} & 31.63	&24.32	&41.50	&46.97&79.85\\
	ByteTrack~\cite{zhang2022bytetrack}&32.12&24.40	&42.33	&50.55	&75.07 \\  
      Ours                                  &\textbf{35.54}&\textbf{28.25}&\textbf{46.25}&\textbf{50.69}&\textbf{80.67} \\
        \hline
        \end{tabular}
        \vspace{-1mm}
        \caption{Comparison on cross-frame association way.  }
    \end{subtable}
        \vspace{2mm}
        \ \ 
    \setlength{\tabcolsep}{5pt}
    \begin{subtable}[t]{0.48\linewidth}
        \begin{tabular}{p{2.6cm}|ccccc}
        \hline\thickhline
        \rowcolor[gray]{0.9}
        Linguistic Extractor& HOTA & DetA & AssA & DetRe & DetPr\\
        \hline
	FastText~\cite{fasttext} 			& 32.39&23.40&46.03&37.03&36.11\\
 Glove~\cite{glove} 				& 32.45&23.71&46.18&39.02&35.03	\\
 Distill-BERT~\cite{distilbert}		& 33.56&26.60&44.01&38.21&43.09\\
  BERT~\cite{bert} 				& 35.28&28.14&45.73&38.32&47.07  \\
    RoBERTa~\cite{liu2019roberta}    & \textbf{35.54}&\textbf{28.25}&\textbf{46.25}&\textbf{39.22}&\textbf{45.94}\\
        \hline
        \end{tabular}
        \vspace{-1mm}
        \caption{Comparison on  linguistic extractor. }
    \end{subtable}
      \vspace{2mm}
    \  \   \   \  
    \setlength{\tabcolsep}{6pt}
    \begin{subtable}[t]{0.48\linewidth}
    \footnotesize
        \begin{tabular}{c|ccccc}
        \hline\thickhline
        \rowcolor[gray]{0.9}
        Referring threshold  & HOTA & DetA & AssA & DetRe & DetPr\\
        \hline
			0.2	&35.10&	26.98&\textbf{	47.15}&	\textbf{41.77	} &39.91 \\
			0.3	&35.07&	27.82&	45.63&	41.74&41.82 \\
			0.4	&\textbf{35.54}&	\textbf{28.25}&	46.25&	39.22&45.94\\
			0.5	&34.73&	26.45&	47.11&	34.16	&49.20\\
			0.6	&31.09&	23.18&	43.27&	27.47	&\textbf{54.07}\\
			0.7	&31.63&	23.20&	44.66&	27.75	&52.84\\
        \hline
        \end{tabular}
        \vspace{-1mm}
        \caption{Comparison on  referring threshold $\beta_{ref}$. }
    \end{subtable}
    \label{tab:array}
      \vspace{-4mm}
  \caption{
    \textbf{Ablation studies of different components} in TransRMOT. HOTA scores are reported, and `-' means unavailable.
    }
         \label{table:ablation}
\end{center}
\vspace{-4pt}
\end{table*}

 \subsection{Qualitative Results}
We visualize several typical referent results in Fig.~\ref{fig:results}.
As seen, TransRMOT is able to detect and track targets accurately under various challenging situations, including multiple objects, status variance,  and varying object numbers.
Besides, we provide a qualitative comparison between all predicted objects and the referent objects from TransRMOT in Fig.~\ref{fig:results_all}. As observed, all visible objects in the video are detected, and the referent objects are also highlighted based on the given expression query.
More qualitative results can be found in \textit{supplementary materials}.

\subsection{Ablation Study}

To study the effect of core components in our model, we conduct extensive ablation studies on Refer-KITTI.

\noindent\textbf{Cross-modal Fusion.}
As described before, the separate early-fusion module is used to model cross-modal representation.
To explore its effect, we remove this module to formulate a new model without expressions as input, which will predict all visible objects.
As shown in Table~\ref{table:ablation} (a), the lack of our cross-modal fusion causes a large performance degradation under all metrics (\eg, HOTA: 34.29$\rightarrow$17.01, DetA: 28.25$\rightarrow$19.05, AssA: 46.25$\rightarrow$15.26).
We also investigate two different variants. As depicted in Table~\ref{table:ablation} (a), the first type is to concatenate and input visual and linguistic features into an encoder, as identical with MDETR~\cite{mdetr}.
The outputted visual part is split and fed into the decoder.
The second type (\ie, language as query) sums up sentence-level language embedding with decoder query to probe the corresponding objects.
Our method achieves better results than both, demonstrating the effectiveness of our early-fusion cross-modal module.


\begin{table}[t]\small{
		\centering
            \footnotesize
		\resizebox{\linewidth}{!}{
			\setlength\tabcolsep{8pt}
			\renewcommand\arraystretch{1.0}
			\begin{tabular}{l|ccccc}
			\hline
            \thickhline
                    \rowcolor[gray]{0.9}
				Ratios & 50\% & 75\%  & 100\%\\ 
				\hline
				Expressions	 & 35.19$\pm$0.68 & 35.48$\pm$0.81 &35.54$\pm$0.71\\ 
                    Videos  & 32.01$\pm$0.82 & 35.44$\pm$0.73 &35.54$\pm$0.71   \\
				\hline 
		\end{tabular} }
		\caption{Ablation studies of the different ratios of expressions or videos on Refer-KITTI in terms of HOTA. }
		\vspace{-12pt}
		\label{table:ratio}}
\end{table}

\noindent\textbf{Cross-frame Association.}
It is also of interest to analyze the impact of cross-frame association using the track query. 
Removing track query (\ie, \textit{w/o} track query) causes TransRMOT to be a fully image-based model.
Table~\ref{table:ablation} (b) shows that it results in numerous IDs and unavailable associations in terms of metrics AssA, AssRe and AssPr.
We associate the referent boxes predicted from the image-based model using state-of-art IoU-matching methods SORT ~\cite{Bewley2016_sort} and ByteTrack~\cite{zhang2022bytetrack}.
Despite achieving association, they have lower HOTA scores than our track query. These experiments approve the necessity of our decoupled query.

\noindent\textbf{Linguistic  Extractor.}
Next, we study different linguistic extractors,  including the widely-used Transformer-based text encoders (\eg, BERT~\cite{bert} and Distill-BERT~\cite{distilbert}) and the simple word embedding methods (\eg, Glove~\cite{glove} and FastText~\cite{fasttext}).
As shown in Table~\ref{table:ablation} (c), these  Transformer-based encoders achieve comparable performance in comparison to the  RoBERTa~\cite{liu2019roberta}, while the simple embedding methods are insufficient in our cross-modal learning.

\noindent\textbf{Referring Threshold.} 
At last, we investigate the effect of referring threshold $\beta_{ref}$. 
As reported in  Table~\ref{table:ablation} (d), the HOTA score is marginal at around 0.2$\sim$0.5 and begins to have a slight reduction when $ \beta _ {ref} $ gets larger. 
Overall, the referring performance is robust to the varying referring threshold.
In this work, we choose $\beta_{ref}\!=\!0.4$ as default.

\begin{figure}[t]
	\centering
	\includegraphics[width=\linewidth]{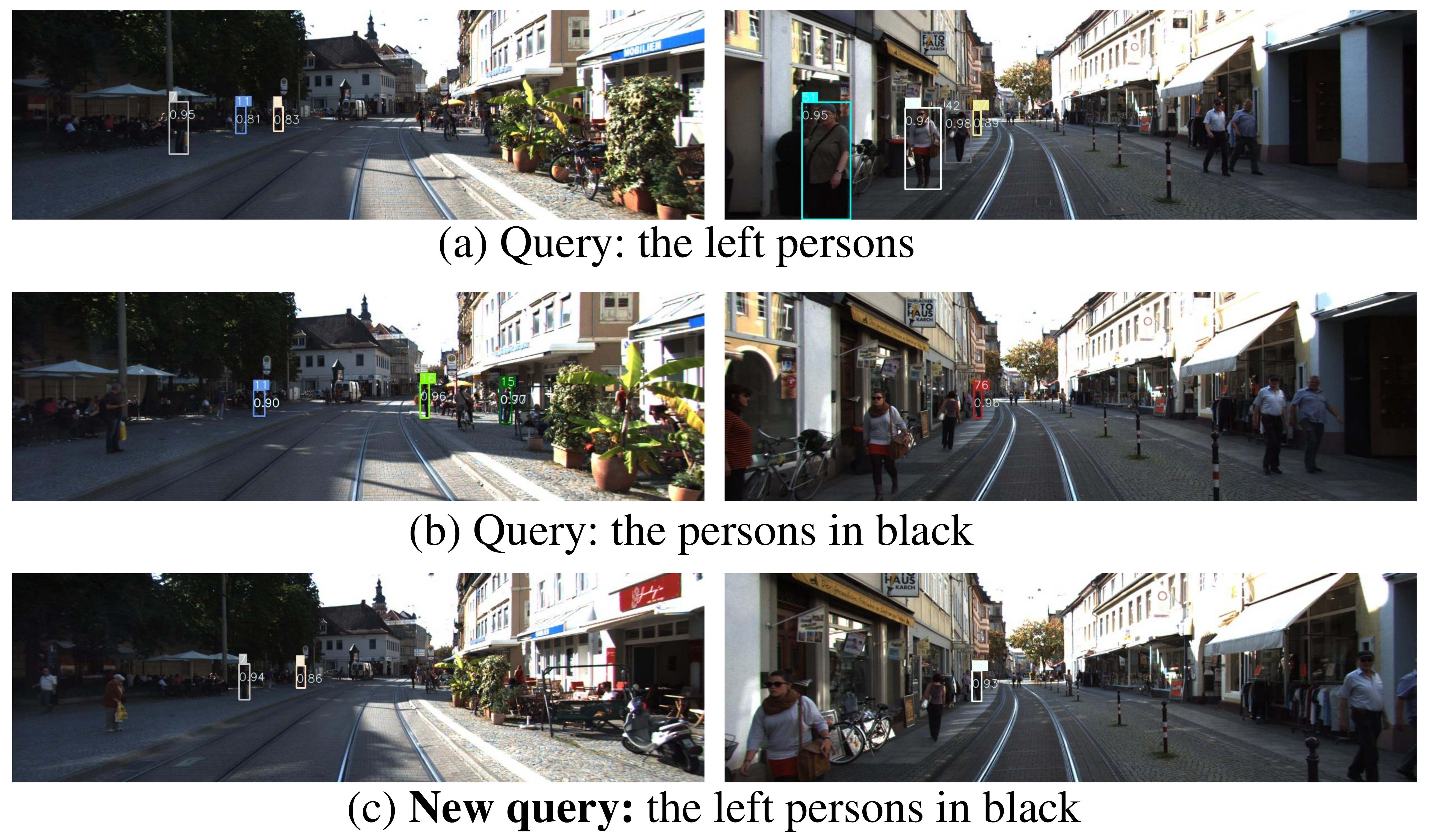}
	\vspace{-20pt}
	\caption{\textbf{Generalization analysis of TransRMOT}. The new query (c) is not included in dataset, but our model can still infer the referent objects according to the existing knowledge (a) (b).}
	\label{fig:general}
\vspace{-2mm}
\end{figure}

\subsection{Generalization Analysis}
\label{section:generalization}
To evaluate the generalization of Refer-KITTI, we train TransRMOT with different ratios of expressions or videos. As shown in Table~\ref{table:ratio},  50\% of expressions already achieve stable HOTA scores, and 75\% of videos lead to similar HOTA performance compared to the full dataset, which shows the dataset itself has enough generalization ability on expressions and videos.
In addition, we directly test on BDD100K~\cite{yu2020bdd100k} with model pretrained on Refer-KITTI (see \href{https://drive.google.com/drive/folders/1wjhM7YJqRCyKJsgY_i1_VvxKjAWgrTN-?usp=sharing}{here} for some examples).
The promising referent results indicate that our dataset also has a good generalization ability across different datasets.


As reported in previous works~\cite{mdetr,huynh2022open,zareian2021open}, language descriptions have a significant advantage in recognition generalization. 
Even if a new expression does not exist in the dataset, the referring understanding model can reason the referent objects by learning existing language knowledge.
To verify this point, we employ a new expression, `the left persons in black', to test TransRMOT. 
Although Refer-KITTI contains some succinct expressions, \eg, `the left persons' and `the persons in black', the new expression is not included in the whole dataset.
In Fig.~\ref{fig:general}, TransRMOT can correctly recognize the referent objects, showing the powerful generalization ability of TransRMOT.

\section{Conclusion}
In this paper, we proposed a novel referring understanding task, called Referring Multi-Object Tracking (RMOT). 
It addressed the single-object limitation of referring understanding tasks and provided a more flexible multi-object setting. Additionally, it leveraged the essential temporal status variant into referring understanding. Both two new settings make RMOT more general, which is appropriate for evaluating real-world requirements.
To promote RMOT, we developed a new benchmark, named Refer-KITTI.  The benchmark provided high flexibility with referent objects and high temporal dynamics but yielded low labeling costs.
Furthermore, we proposed a Transformer-based method, TransRMOT, to tackle the new task.
The framework is fully end-to-end optimized during training, and predicts referent objects frame by frame.
We validated TransRMOT on Refer-KITTI, and it achieved state-of-art performance.




\section{Supplementary Material}

\subsection{Competitor Details}
\begin{figure}[h]
	\centering
	\includegraphics[width=\linewidth]{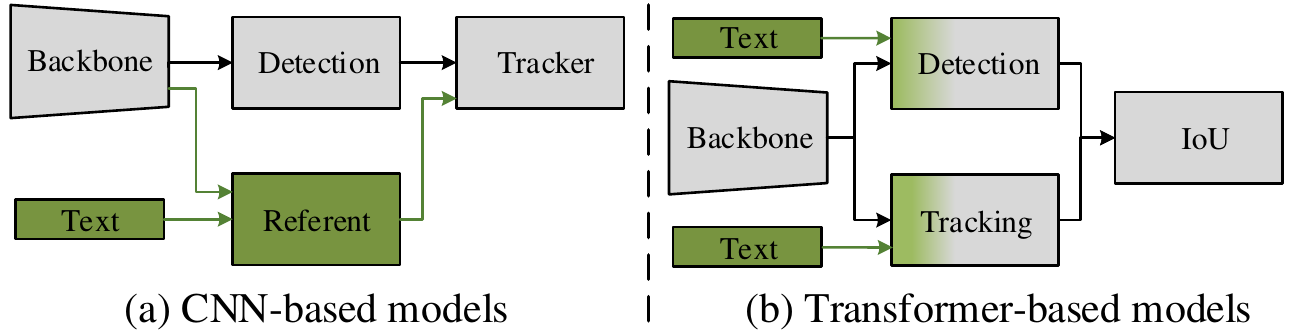}
	\caption{The \textcolor{gray}{original MOT models} and our \textcolor{ForestGreen}{cross-modal modification} for CNN-based and Transformer-based framework.}
	\label{fig:model_supp}
	\vspace{-4pt}
\end{figure}

We provide more model details about the competitors (introduced in \S~\ref{sec:quantitative_results}).
The CNN-based counterparts build upon several multi-object tracking (MOT) models, such as  FairMOT~\cite{zhang2021fairmot}, DeepSORT~\cite{Wojke2017simple}, ByteTrack~\cite{zhang2022bytetrack},  and CStrack~\cite{liang2022rethinking},  with some crucial modifications on cross-modal learning.
In specific,  these CNN-based MOT models typically follow a \textit{tracking-by-detection} pattern, which consists of a detector (including backbone and detection head) for single-frame detection  and a tracker for cross-frame object association.
As shown in Fig.~\ref{fig:model_supp}(a), we design a \textcolor{ForestGreen}{referent branch} on the visual backbone.
It contains our proposed cross-modal fusion module and the detection head from the original MOT model.
The cross-modal module fuses visual and linguistic features and provides comprehensive feature representation. The detection head decodes the fused feature maps into object boxes with the same format as the original outputs.
During training, we keep the losses of predicting all visible objects.
For inference, the  default tracker is used to associate cross-frame referent objects.
DeepSORT and ByteTrack do not provide a detection model, so we employ the referent results from FairMOT.

In addition to CNN-based methods, we also experiment with Transformer-based 
MOT models,  such as TransTrack~\cite{sun2020transtrack} and TrackFormer~\cite{meinhardt2022trackformer}. 
We modify them by adding our cross-modal early-fusion module before the encoder layers, as depicted in Fig.~\ref{fig:model_supp}(b).

\subsection{Limitation}
\begin{figure}[h]
	\centering
	\includegraphics[width=\linewidth]{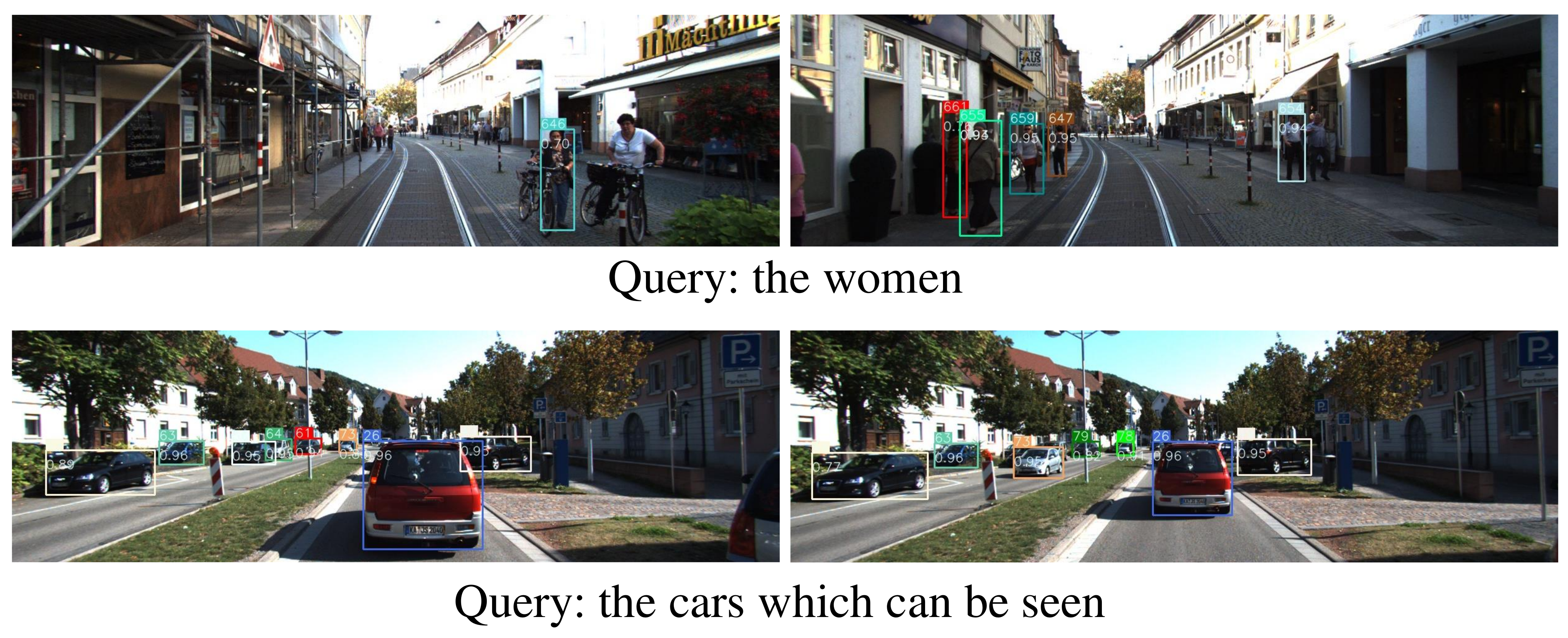}
	\vspace{-22pt}
	\caption{Typical failure cases from TransRMOT.}
	\label{fig:results_more}
	\vspace{-12pt}
    \label{fig:failure_cases}
\end{figure}

Fig.~\ref{fig:failure_cases} visualizes several failure cases from TransRMOT. 
The first case is that some fine-grained object features (\eg, human gender) are not captured accurately, hindering the detection performance.
To avoid this case, the top-down solution (\ie, the detection-then-fusion method) can be jointly explored to focus more on the fine-grained features of object regions.
The second case is ID switch problem, which is caused by  long-temporal occlusion  and degrades the tracking performance. 
To address this problem, object representation can keep more time for long-term association using a memory mechanism in future work.

\subsection{More Qualitative Results}

We offer more qualitative results in Fig.~\ref{fig:results_more}. As seen, our proposed TransRMOT achieves compelling results under various challenging situations, \eg, multiple objects,  object entrance and exit, moving objects, occlusion and \etc.

\begin{figure*}[t]
	\centering
	\includegraphics[width=\linewidth]{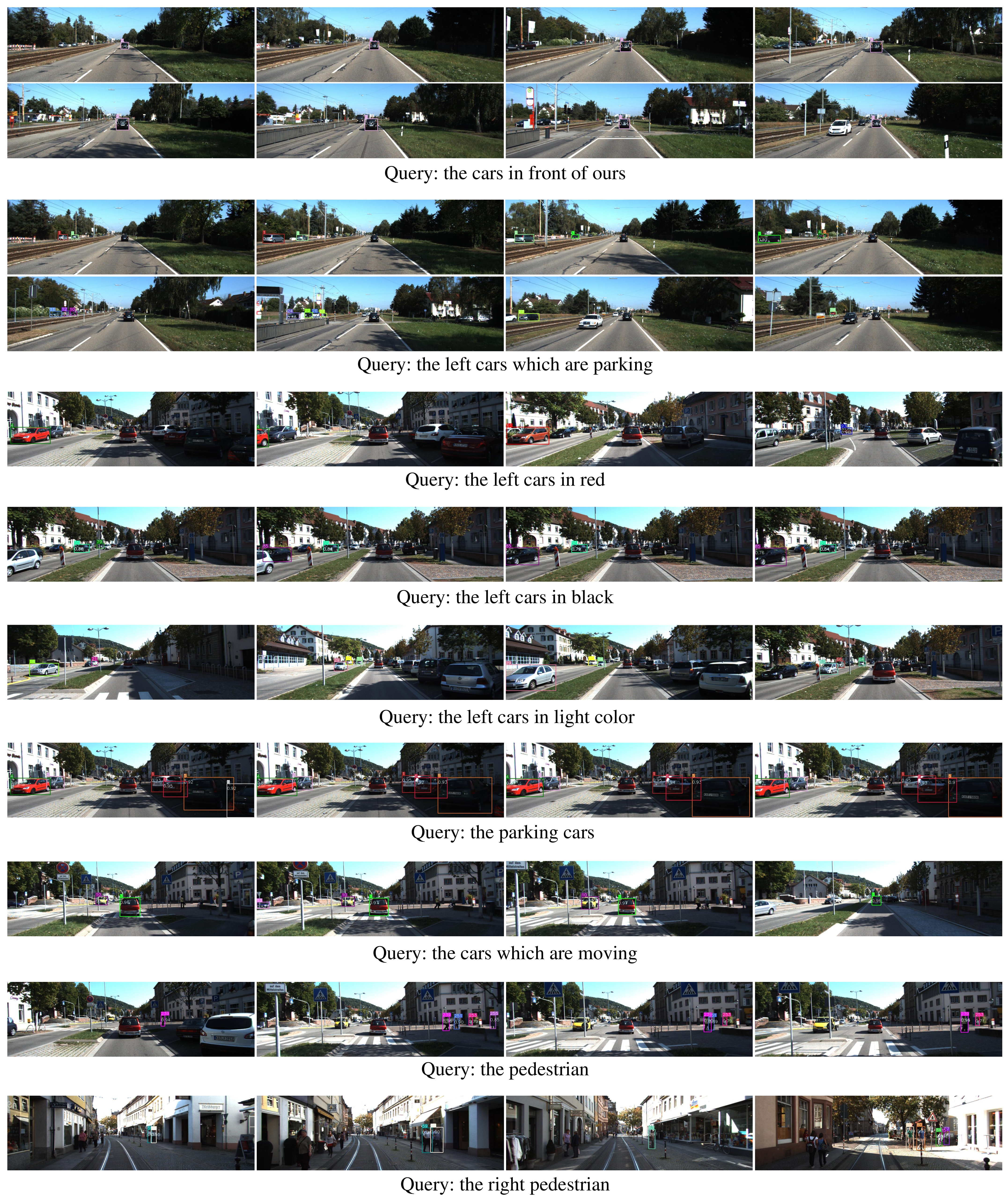}
	\vspace{-10pt}
	\caption{More qualitative results on Refer-KITTI.}
	\label{fig:results_more}
	\vspace{-4pt}
\end{figure*}

{\small
\bibliographystyle{ieee_fullname}
\bibliography{egbib}

\begin{thebibliography}{10}\itemsep=-1pt

\bibitem{Bewley2016_sort}
Alex Bewley, Zongyuan Ge, Lionel Ott, Fabio Ramos, and Ben Upcroft.
\newblock Simple online and realtime tracking.
\newblock In {\em ICIP}, 2016.

\bibitem{mttr}
Adam Botach, Evgenii Zheltonozhskii, and Chaim Baskin.
\newblock End-to-end referring video object segmentation with multimodal
  transformers.
\newblock In {\em CVPR}, 2022.

\bibitem{carion2020end}
Nicolas Carion, Francisco Massa, Gabriel Synnaeve, Nicolas Usunier, Alexander
  Kirillov, and Sergey Zagoruyko.
\newblock End-to-end object detection with transformers.
\newblock In {\em ECCV}, 2020.

\bibitem{chen2019weakly}
Zhenfang Chen, Lin Ma, Wenhan Luo, and Kwan-Yee~K Wong.
\newblock Weakly-supervised spatio-temporally grounding natural sentence in
  video.
\newblock In {\em ACL}, 2019.

\bibitem{talk2car}
Thierry Deruyttere, Simon Vandenhende, Dusan Grujicic, Luc Van~Gool, and
  Marie-Francine Moens.
\newblock Talk2car: Taking control of your self-driving car.
\newblock {\em arXiv preprint arXiv:1909.10838}, 2019.

\bibitem{bert}
Jacob Devlin, Ming-Wei Chang, Kenton Lee, and Kristina Toutanova.
\newblock Bert: Pre-training of deep bidirectional transformers for language
  understanding.
\newblock {\em arXiv preprint arXiv:1810.04805}, 2018.

\bibitem{ding2022language}
Zihan Ding, Tianrui Hui, Junshi Huang, Xiaoming Wei, Jizhong Han, and Si Liu.
\newblock Language-bridged spatial-temporal interaction for referring video
  object segmentation.
\newblock In {\em CVPR}, 2022.

\bibitem{gavrilyuk2018actor}
Kirill Gavrilyuk, Amir Ghodrati, Zhenyang Li, and Cees~GM Snoek.
\newblock Actor and action video segmentation from a sentence.
\newblock In {\em CVPR}, 2018.

\bibitem{Geiger2012CVPR}
Andreas Geiger, Philip Lenz, and Raquel Urtasun.
\newblock Are we ready for autonomous driving? the kitti vision benchmark
  suite.
\newblock In {\em CVPR}, 2012.

\bibitem{he2016deep}
Kaiming He, Xiangyu Zhang, Shaoqing Ren, and Jian Sun.
\newblock Deep residual learning for image recognition.
\newblock In {\em CVPR}, 2016.

\bibitem{huang2020referring}
Shaofei Huang, Tianrui Hui, Si Liu, Guanbin Li, Yunchao Wei, Jizhong Han, Luoqi
  Liu, and Bo Li.
\newblock Referring image segmentation via cross-modal progressive
  comprehension.
\newblock In {\em CVPR}, 2020.

\bibitem{hui2021collaborative}
Tianrui Hui, Shaofei Huang, Si Liu, Zihan Ding, Guanbin Li, Wenguan Wang,
  Jizhong Han, and Fei Wang.
\newblock Collaborative spatial-temporal modeling for language-queried video
  actor segmentation.
\newblock In {\em CVPR}, 2021.

\bibitem{huynh2022open}
Dat Huynh, Jason Kuen, Zhe Lin, Jiuxiang Gu, and Ehsan Elhamifar.
\newblock Open-vocabulary instance segmentation via robust cross-modal
  pseudo-labeling.
\newblock In {\em CVPR}, 2022.

\bibitem{mdetr}
Aishwarya Kamath, Mannat Singh, Yann LeCun, Gabriel Synnaeve, Ishan Misra, and
  Nicolas Carion.
\newblock Mdetr-modulated detection for end-to-end multi-modal understanding.
\newblock In {\em ICCV}, 2021.

\bibitem{kazemzadeh2014referitgame}
Sahar Kazemzadeh, Vicente Ordonez, Mark Matten, and Tamara Berg.
\newblock Referitgame: Referring to objects in photographs of natural scenes.
\newblock In {\em EMNLP}, 2014.

\bibitem{kesen2022modulating}
Ilker Kesen, Ozan~Arkan Can, Erkut Erdem, Aykut Erdem, and Deniz Y{\"u}ret.
\newblock Modulating bottom-up and top-down visual processing via
  language-conditional filters.
\newblock In {\em CVPR}, 2022.

\bibitem{khoreva2018video}
Anna Khoreva, Anna Rohrbach, and Bernt Schiele.
\newblock Video object segmentation with language referring expressions.
\newblock In {\em ACCV}, 2018.

\bibitem{li2021referring}
Muchen Li and Leonid Sigal.
\newblock Referring transformer: A one-step approach to multi-task visual
  grounding.
\newblock {\em NeurIPS}, 2021.

\bibitem{li2017tracking}
Zhenyang Li, Ran Tao, Efstratios Gavves, Cees~GM Snoek, and Arnold~WM
  Smeulders.
\newblock Tracking by natural language specification.
\newblock In {\em CVPR}, 2017.

\bibitem{liang2022rethinking}
Chao Liang, Zhipeng Zhang, Xue Zhou, Bing Li, Shuyuan Zhu, and Weiming Hu.
\newblock Rethinking the competition between detection and reid in multiobject
  tracking.
\newblock {\em IEEE TIP}, 2022.

\bibitem{liao2020real}
Yue Liao, Si Liu, Guanbin Li, Fei Wang, Yanjie Chen, Chen Qian, and Bo Li.
\newblock A real-time cross-modality correlation filtering method for referring
  expression comprehension.
\newblock In {\em CVPR}, 2020.

\bibitem{lin2017focal}
Tsung-Yi Lin, Priya Goyal, Ross Girshick, Kaiming He, and Piotr Doll{\'a}r.
\newblock Focal loss for dense object detection.
\newblock In {\em ICCV}, 2017.

\bibitem{coco}
Tsung-Yi Lin, Michael Maire, Serge Belongie, James Hays, Pietro Perona, Deva
  Ramanan, Piotr Doll{\'a}r, and C~Lawrence Zitnick.
\newblock Microsoft coco: Common objects in context.
\newblock In {\em ECCV}, 2014.

\bibitem{liu2019learning}
Daqing Liu, Hanwang Zhang, Feng Wu, and Zheng-Jun Zha.
\newblock Learning to assemble neural module tree networks for visual
  grounding.
\newblock In {\em ICCV}, 2019.

\bibitem{liu2019roberta}
Yinhan Liu, Myle Ott, Naman Goyal, Jingfei Du, Mandar Joshi, Danqi Chen, Omer
  Levy, Mike Lewis, Luke Zettlemoyer, and Veselin Stoyanov.
\newblock Roberta: A robustly optimized bert pretraining approach.
\newblock {\em arXiv preprint arXiv:1907.11692}, 2019.

\bibitem{luiten2020IJCV}
Jonathon Luiten, Aljosa Osep, Patrick Dendorfer, Philip Torr, Andreas Geiger,
  Laura Leal-Taix{\'e}, and Bastian Leibe.
\newblock Hota: A higher order metric for evaluating multi-object tracking.
\newblock {\em IJCV}, 2020.

\bibitem{luo2020multi}
Gen Luo, Yiyi Zhou, Xiaoshuai Sun, Liujuan Cao, Chenglin Wu, Cheng Deng, and
  Rongrong Ji.
\newblock Multi-task collaborative network for joint referring expression
  comprehension and segmentation.
\newblock In {\em CVPR}, 2020.

\bibitem{luo2017comprehension}
Ruotian Luo and Gregory Shakhnarovich.
\newblock Comprehension-guided referring expressions.
\newblock In {\em CVPR}, 2017.

\bibitem{mao2016generation}
Junhua Mao, Jonathan Huang, Alexander Toshev, Oana Camburu, Alan Yuille, and
  Kevin Murphy.
\newblock Generation and comprehension of unambiguous object descriptions.
\newblock In {\em CVPR}, 2016.

\bibitem{meinhardt2022trackformer}
Tim Meinhardt, Alexander Kirillov, Laura Leal-Taixe, and Christoph
  Feichtenhofer.
\newblock Trackformer: Multi-object tracking with transformers.
\newblock In {\em CVPR}, 2022.

\bibitem{fasttext}
Tomas Mikolov, Edouard Grave, Piotr Bojanowski, Christian Puhrsch, and Armand
  Joulin.
\newblock Advances in pre-training distributed word representations.
\newblock In {\em LREC}, 2018.

\bibitem{nagaraja2016modeling}
Varun~K Nagaraja, Vlad~I Morariu, and Larry~S Davis.
\newblock Modeling context between objects for referring expression
  understanding.
\newblock In {\em ECCV}, 2016.

\bibitem{glove}
Jeffrey Pennington, Richard Socher, and Christopher~D Manning.
\newblock Glove: Global vectors for word representation.
\newblock In {\em EMNLP}, 2014.

\bibitem{giou}
Hamid Rezatofighi, Nathan Tsoi, JunYoung Gwak, Amir Sadeghian, Ian Reid, and
  Silvio Savarese.
\newblock Generalized intersection over union: A metric and a loss for bounding
  box regression.
\newblock In {\em CVPR}, 2019.

\bibitem{rufus2020cosine}
Nivedita Rufus, Unni Krishnan~R Nair, K~Madhava Krishna, and Vineet Gandhi.
\newblock Cosine meets softmax: A tough-to-beat baseline for visual grounding.
\newblock In {\em ECCV}, 2020.

\bibitem{sadhu2020video}
Arka Sadhu, Kan Chen, and Ram Nevatia.
\newblock Video object grounding using semantic roles in language description.
\newblock In {\em CVPR}, 2020.

\bibitem{distilbert}
Victor Sanh, Lysandre Debut, Julien Chaumond, and Thomas Wolf.
\newblock Distilbert, a distilled version of bert: smaller, faster, cheaper and
  lighter.
\newblock {\em arXiv preprint arXiv:1910.01108}, 2019.

\bibitem{seo2020urvos}
Seonguk Seo, Joon-Young Lee, and Bohyung Han.
\newblock Urvos: Unified referring video object segmentation network with a
  large-scale benchmark.
\newblock In {\em ECCV}, 2020.

\bibitem{song2021co}
Sijie Song, Xudong Lin, Jiaying Liu, Zongming Guo, and Shih-Fu Chang.
\newblock Co-grounding networks with semantic attention for referring
  expression comprehension in videos.
\newblock In {\em CVPR}, 2021.

\bibitem{sun2020transtrack}
Peize Sun, Jinkun Cao, Yi Jiang, Rufeng Zhang, Enze Xie, Zehuan Yuan, Changhu
  Wang, and Ping Luo.
\newblock Transtrack: Multiple object tracking with transformer.
\newblock {\em arXiv preprint arXiv:2012.15460}, 2020.

\bibitem{tan2021look}
Reuben Tan, Bryan Plummer, Kate Saenko, Hailin Jin, and Bryan Russell.
\newblock Look at what i’m doing: Self-supervised spatial grounding of
  narrations in instructional videos.
\newblock {\em NeurIPS}, 2021.

\bibitem{vasudevan2018object}
Arun~Balajee Vasudevan, Dengxin Dai, and Luc Van~Gool.
\newblock Object referring in videos with language and human gaze.
\newblock In {\em CVPR}, 2018.

\bibitem{vaswani2017attention}
Ashish Vaswani, Noam Shazeer, Niki Parmar, Jakob Uszkoreit, Llion Jones,
  Aidan~N Gomez, {\L}ukasz Kaiser, and Illia Polosukhin.
\newblock Attention is all you need.
\newblock {\em NeurIPS}, 2017.

\bibitem{wang2019neighbourhood}
Peng Wang, Qi Wu, Jiewei Cao, Chunhua Shen, Lianli Gao, and Anton van~den
  Hengel.
\newblock Neighbourhood watch: Referring expression comprehension via
  language-guided graph attention networks.
\newblock In {\em CVPR}, 2019.

\bibitem{Wojke2017simple}
Nicolai Wojke, Alex Bewley, and Dietrich Paulus.
\newblock Simple online and realtime tracking with a deep association metric.
\newblock In {\em ICIP}, 2017.

\bibitem{wu2022multi}
Dongming Wu, Xingping Dong, Ling Shao, and Jianbing Shen.
\newblock Multi-level representation learning with semantic alignment for
  referring video object segmentation.
\newblock In {\em CVPR}, 2022.

\bibitem{referformer}
Jiannan Wu, Yi Jiang, Peize Sun, Zehuan Yuan, and Ping Luo.
\newblock Language as queries for referring video object segmentation.
\newblock In {\em CVPR}, 2022.

\bibitem{yang2022tubedetr}
Antoine Yang, Antoine Miech, Josef Sivic, Ivan Laptev, and Cordelia Schmid.
\newblock Tubedetr: Spatio-temporal video grounding with transformers.
\newblock In {\em CVPR}, 2022.

\bibitem{yang2019cross}
Sibei Yang, Guanbin Li, and Yizhou Yu.
\newblock Cross-modal relationship inference for grounding referring
  expressions.
\newblock In {\em CVPR}, 2019.

\bibitem{yang2020graph}
Sibei Yang, Guanbin Li, and Yizhou Yu.
\newblock Graph-structured referring expression reasoning in the wild.
\newblock In {\em CVPR}, 2020.

\bibitem{yang2021bottom}
Sibei Yang, Meng Xia, Guanbin Li, Hong-Yu Zhou, and Yizhou Yu.
\newblock Bottom-up shift and reasoning for referring image segmentation.
\newblock In {\em CVPR}, 2021.

\bibitem{ye2019cross}
Linwei Ye, Mrigank Rochan, Zhi Liu, and Yang Wang.
\newblock Cross-modal self-attention network for referring image segmentation.
\newblock In {\em CVPR}, 2019.

\bibitem{young2014image}
Peter Young, Alice Lai, Micah Hodosh, and Julia Hockenmaier.
\newblock From image descriptions to visual denotations: New similarity metrics
  for semantic inference over event descriptions.
\newblock {\em TACL}, 2014.

\bibitem{yu2020bdd100k}
Fisher Yu, Haofeng Chen, Xin Wang, Wenqi Xian, Yingying Chen, Fangchen Liu,
  Vashisht Madhavan, and Trevor Darrell.
\newblock Bdd100k: A diverse driving dataset for heterogeneous multitask
  learning.
\newblock In {\em CVPR}, 2020.

\bibitem{yu2018mattnet}
Licheng Yu, Zhe Lin, Xiaohui Shen, Jimei Yang, Xin Lu, Mohit Bansal, and
  Tamara~L Berg.
\newblock Mattnet: Modular attention network for referring expression
  comprehension.
\newblock In {\em CVPR}, 2018.

\bibitem{yu2016modeling}
Licheng Yu, Patrick Poirson, Shan Yang, Alexander~C Berg, and Tamara~L Berg.
\newblock Modeling context in referring expressions.
\newblock In {\em ECCV}, 2016.

\bibitem{zareian2021open}
Alireza Zareian, Kevin~Dela Rosa, Derek~Hao Hu, and Shih-Fu Chang.
\newblock Open-vocabulary object detection using captions.
\newblock In {\em CVPR}, 2021.

\bibitem{zeng2021motr}
Fangao Zeng, Bin Dong, Tiancai Wang, Xiangyu Zhang, and Yichen Wei.
\newblock Motr: End-to-end multiple-object tracking with transformer.
\newblock In {\em ECCV}, 2021.

\bibitem{zhang2022bytetrack}
Yifu Zhang, Peize Sun, Yi Jiang, Dongdong Yu, Fucheng Weng, Zehuan Yuan, Ping
  Luo, Wenyu Liu, and Xinggang Wang.
\newblock Bytetrack: Multi-object tracking by associating every detection box.
\newblock In {\em ECCV}, 2022.

\bibitem{zhang2021fairmot}
Yifu Zhang, Chunyu Wang, Xinggang Wang, Wenjun Zeng, and Wenyu Liu.
\newblock Fairmot: On the fairness of detection and re-identification in
  multiple object tracking.
\newblock {\em IJCV}, 2021.

\bibitem{zhang2022motrv2}
Yuang Zhang, Tiancai Wang, and Xiangyu Zhang.
\newblock Motrv2: Bootstrapping end-to-end multi-object tracking by pretrained
  object detectors.
\newblock In {\em CVPR}, 2023.

\bibitem{zhu2020deformable}
Xizhou Zhu, Weijie Su, Lewei Lu, Bin Li, Xiaogang Wang, and Jifeng Dai.
\newblock Deformable detr: Deformable transformers for end-to-end object
  detection.
\newblock In {\em ICLR}, 2020.

\end{thebibliography}
}

\end{document}